\documentclass[10pt,twocolumn,letterpaper]{article}

\usepackage[pagenumbers]{cvpr} 

\usepackage{multicol}
\newcommand{\customtitle}{
  \begin{center}
    {\LARGE \bfseries Supplementary Material for LaMoGen: Language to Motion Generation Through
    LLM-Guided Symbolic Inference}
    \vspace{1em}
    
    {\large
      Junkun Jiang, Ho Yin Au, Jingyu Xiang, Jie Chen\footnotemark[1] \\[0.5em]
      Department of Computer Science, Hong Kong Baptist University, HKSAR \\[0.3em]
      {\tt\small \{csjkjiang, cshyau, csjyxiang, chenjie\}@comp.hkbu.edu.hk}
    }
    
    \vspace{0.8em}
  \end{center}
}

\usepackage{url}
\usepackage{graphicx}
\usepackage{comment}
\usepackage{booktabs}
\usepackage{multirow}
\usepackage{arydshln}
\usepackage{amssymb}
\usepackage[table]{xcolor}
\usepackage{natbib}
\usepackage{multirow} 
\usepackage{tabularx}
\usepackage[table]{xcolor}
\usepackage{longtable}
\usepackage{pifont}
\newcommand{\cmark}{\ding{51}}%
\newcommand{\xmark}{\ding{55}}%
\usepackage{wrapfig,lipsum,booktabs}
\definecolor{fst}{RGB}{255,235,205}
\definecolor{sec}{RGB}{220,235,255}
\definecolor{trd}{RGB}{220,245,220}

\definecolor{cvprblue}{rgb}{0.21,0.49,0.74}
\usepackage[pagebackref,breaklinks,colorlinks,allcolors=cvprblue]{hyperref}

\def\paperID{8706} 
\def\confName{CVPR}
\def\confYear{2026}

\title{LaMoGen: Language to Motion Generation Through \\ LLM-Guided Symbolic Inference}

\author{Junkun Jiang, Ho Yin Au, Jingyu Xiang, Jie Chen
\thanks{
\parbox[t]{0.8\textwidth}{Corresponding author. \\ Project page: \url{https://jjkislele.github.io/LaMoGen/}}
} \\
Department of Computer Science, Hong Kong Baptist University, HKSAR \\
{\tt\small \{csjkjiang, cshyau, csjyxiang, chenjie\}@comp.hkbu.edu.hk}
}

\begin{document}

\maketitle

\begin{abstract}
Human motion is highly expressive and naturally aligned with language, yet prevailing methods relying heavily on joint text-motion embeddings struggle to synthesize temporally accurate, detailed motions and often lack explainability.
To address these limitations, we introduce LabanLite, a motion representation developed by adapting and extending the Labanotation system. Unlike black-box text–motion embeddings, LabanLite encodes each atomic body-part action (e.g., a single left-foot step) as a discrete Laban symbol paired with a textual template. This abstraction decomposes complex motions into interpretable symbol sequences and body-part instructions, establishing a symbolic link between high-level language and low-level motion trajectories.
Building on LabanLite, we present LaMoGen, a Text-to-\textbf{La}banLite-to-\textbf{Mo}tion \textbf{Gen}eration framework that enables large language models (LLMs) to compose motion sequences through symbolic reasoning. The LLM interprets motion patterns, relates them to textual descriptions, and recombines symbols into executable plans, producing motions that are both interpretable and linguistically grounded.
To support rigorous evaluation, we introduce a Labanotation-based benchmark with structured description–motion pairs and three metrics that jointly measure text–motion alignment across symbolic, temporal, and harmony dimensions.
Experiments demonstrate that LaMoGen establishes a new baseline for both interpretability and controllability, outperforming prior methods on our benchmark and two public datasets. These results highlight the advantages of symbolic reasoning and agent-based design for language-driven motion synthesis. 
\end{abstract}

\section{Introduction}
\begin{figure*}[t]
\centering
\includegraphics[width=0.93\linewidth]{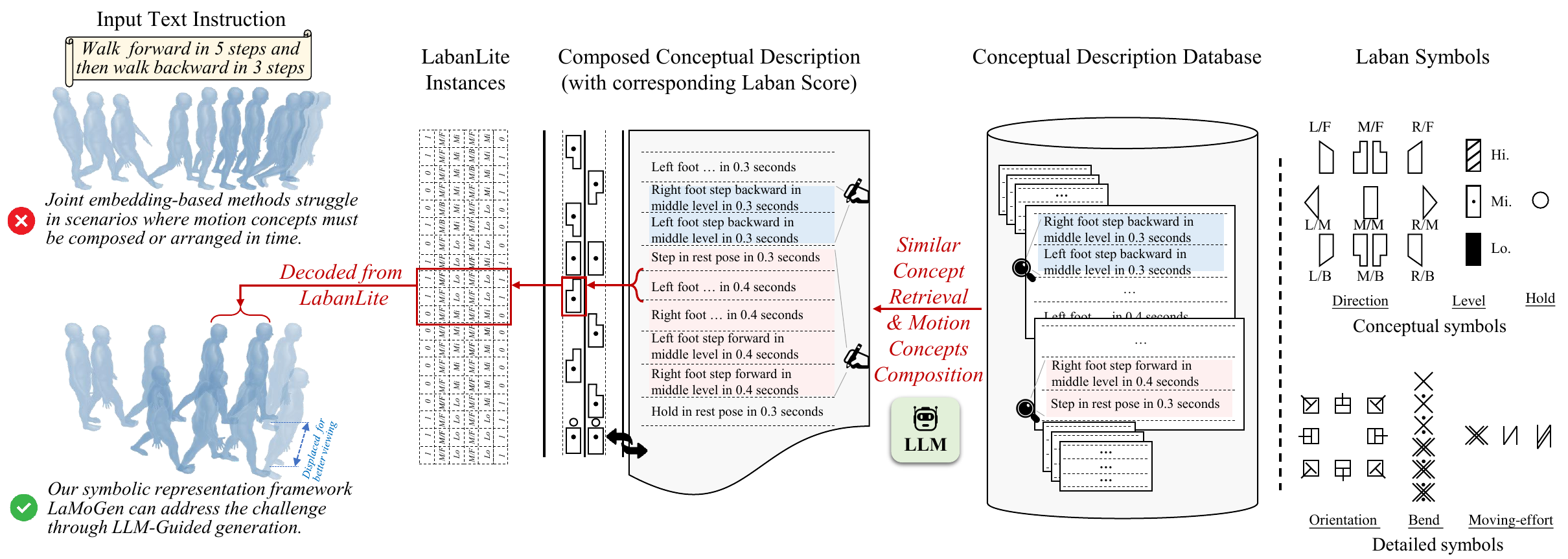}
\vspace{-3.5mm}
\caption{Given a structured text description, methods based on text-motion joint embeddings often fail to generate semantically consistent motion. In contrast, our approach leverages symbolic motion representations, allowing for accurate motion generation. As each symbol is associated with one conceptual description, this design enables LLMs to compose symbolic motion via retrieval augmentation prompting. }
\vspace{-5mm}
\label{fig:teaser}
\end{figure*}

Human motions convey rich semantics that often correspond to intentions and instructions expressed in natural language. Establishing a precise mapping between language and motion is therefore essential for computational understanding and modelling of human behaviour. Despite recent approaches~\citep{guo2022humanml3d,tevet2023mdm,zhang2023t2mgpt,jiang2024motiongpt,au2025deep,au2025soscontrol} have made promising progress in generating motion from textual descriptions, these methods often struggle to capture fine-grained semantics, limiting the fidelity and precision with which generated motions adhere to the intended instructions.
Recent works~\citep{huang2024controllable,li2024lamp,jiang2024motiongpt} attempt to address this issue by decomposing text into body-part-level tokens for more precise spatial pose specification. However, such representations do not easily generalize to the temporal domain. For example, as shown in Fig.~\ref{fig:teaser}, instructions like ``Walk forward in 5 steps and then walk backward in 3 steps'' are often interpreted as a generic sequence of ``step forward'' motions, failing to accurately reflect the specified number of steps and the temporal order of actions. 

On the other hand, the traditional symbolic annotation system, Labanotation~\cite{topaz1996elementary}, encodes detailed features of movement—such as specific body parts, direction, level, duration, and other qualitative attributes—into interpretable symbols. This allows for a compact, precise, and prescriptive representation of motion, facilitating clearer specification of actions described in input texts.
For example, the instruction ``Walk forward in 5 steps and then walk backward in 3 steps'' can be efficiently decomposed into a sequence of symbols representing each atomic movement, as shown in the score from Fig.~\ref{fig:teaser}. 
More importantly, by coupling each symbol with a fixed textual description, motions and texts can be converted back and forth unambiguously. \textit{This enables humans to directly alter movements by editing the symbols, or large language models to modify actions by editing the corresponding text.} Thus, this symbolic system acts as an effective interface between language and motion, clarifying both temporal and spatial action structures.

However, by encapsulating body part involvement and duration as attribute fields, Labanotation introduces a level of abstraction that challenges machine learning models to directly correlate these symbolic representations with continuous motion data.
To address these challenges, we propose \textit{LabanLite}: a frame- and body-part-wise motion representation that explicitly captures the temporal and structural attributes of Labanotation. This not only enhances compatibility with machine learning methods, but also serves as an effective intermediate representation for large language models in motion planning, enabling them to actively interpret and disambiguate complex textual instructions into a structured LabanLite script, thus achieving fine-grained semantic fidelity and precision in generated motions.

Building on LabanLite, we propose   \textit{LaMoGen}—an end-to-end Text-to-\textbf{La}banotation-to-\textbf{Mo}tion \textbf{Gen}eration framework for synthesizing fine-grained motion sequences aligned with complex text instructions.
As shown in Fig.~\ref{fig:teaser}, LaMoGen first maps complex instructions to LabanLite representations via LLM-based planning, and subsequently decodes them into motion sequences. This separation of high-level semantic planning from low-level motion synthesis enables LaMoGen to generate interpretable motions that faithfully represent instruction intent.
Furthermore, by leveraging the interpretability of LabanLite, we propose a Labanotation-based benchmark featuring three metrics that explicitly evaluate motion–text alignment across symbolic, temporal, and harmony dimensions. This benchmark assesses the accuracy of orientation, duration, and coordination for the four main body parts (right/left arms and legs), providing a comprehensive framework for transparent and interpretable evaluation of how well generated motions reflect complex textual instructions.

We evaluate LaMoGen on our Laban benchmark as well as two widely used text-to-motion datasets. Experimental results show that our framework achieves state-of-the-art performance in aligning generated motions with textual descriptions, highlighting the benefits of Labanotation-based representations for both interpretability and controllability in motion synthesis.
To the best of our knowledge, LaMoGen is the first framework to enable LLMs to autonomously compose motion through interpretable symbolic representations.
The main contributions of this work can be summarised as follows:
\begin{itemize}
\item We propose LabanLite, a frame- and body-part-wise motion representation grounded in Labanotation that provides explicit temporal and structural annotation for machine learning and interpretable LLM-based planning.
\item We propose LaMoGen, a two-level framework that unifies high-level LLM-driven symbolic planning with low-level motion synthesis, enabling greater interpretability and controllability compared to prior approaches.
\item We present a Labanotation-based benchmark with novel metrics and a structured dataset to rigorously evaluate motion-text alignment across symbolic, temporal, and harmony dimensions.
\end{itemize}

\section{Related Works}

\noindent\textbf{Text-based Human Motion Generation} aims to generate diverse, human-like motion from natural language descriptions or action labels, typically by mapping paired text and motion data into a joint embedding space~\citep{guo2022humanml3d,tevet2023mdm,lee2023multiact,zhang2024motiondiffuse,li2024lamp,jiang2026mmdm}. Recent work has aimed to abstract motion signals for semantically consistent representations.
Kinematic Phrase (KP)~\citep{liu2024bridging} heuristically abstracts motions via inter-body part distances and uses a variational autoencoder to generate motions. However, such methods are limited to low-level signal abstraction and have difficulty capturing high-level semantic meaning.
On the other hand, Vector Quantised Variational Autoencoders~\citep{zhang2023t2mgpt,jiang2024motiongpt,li2024lamp,zeng2025light,guo2024momask} have been widely adopted to represent motion as discretised tokens, enabling coherent sequence generation with autoregressive transformers.
CoMo~\citep{huang2024controllable} advances this line of work by discretising motion into pose codes using Posescript~\citep{delmas2024posescript}. However, pose codes represent only the key pose state of individual frames, thus lacking the ability to capture transitional dynamics. To address this limitation, we draw inspiration from CoMo and propose the use of Laban symbols as the intermediate representation between text and motion. Unlike pose codes, each Laban symbol encapsulates not only the starting and ending poses but also the transformation process between them. This makes Laban symbols more abstract and semantically expressive. Consequently, our proposed LaMoGen framework enables LLMs to compose Laban symbols directly, as they are more closely aligned with textual representations and facilitate symbolic reasoning about motion.

\noindent\textbf{Controllable Motion Generation}. Several methods introduce explicit signals to guide motion generation, such as using key joint trajectories~\citep{karunratanakul2023gmd,wan2024tlcontrol} or drawing key poses~\citep{liu2023plan,wang2025stickmotion,au2025soscontrol}. However, these approaches require manual input and specialised user interfaces, limiting their intuitiveness and ease of use. In contrast, Labanotation offers a more convenient symbolic representation and thus holds greater potential for controllable motion generation.

\noindent\textbf{Motion Generation with LLMs}.
To our knowledge, no existing work explicitly enables LLMs to autonomously generate motion without fine-tuning~\citep{kalakonda2023action,zhang2024motiongpt,zhou2024avatargpt,jiang2024motiongpt}. The closest prior works~\citep{shi2023generating,athanasiou2023sinc,huang2024controllable} use LLMs to decompose text into body part descriptions, without allowing them to compose or generate motions. In contrast, we adopt a retrieval-augmented prompting strategy, providing LLMs with reference Laban scores to help them understand how to produce the desired motions.

\noindent\textbf{Labanotation}. 
Recent studies primarily use Labanotation as a motion representation for reconstruction tasks. \citet{jiang2024motion} used Laban symbols to explicitly represent and reconstruct inbetweening motions. \citet{li2024translating} mapped hand images to Laban symbols and then to hand motions, to improve hand pose estimation accuracy. In this work, we apply Labanotation to motion generation, bridging textual descriptions and motions. 
To assess the accuracy of Laban symbol detection, we utilize Laban-annotated motion data~\citep{jiang2024motion} and report the corresponding performance.

\begin{figure*}[th]
\centering
\includegraphics[width=0.83\linewidth]{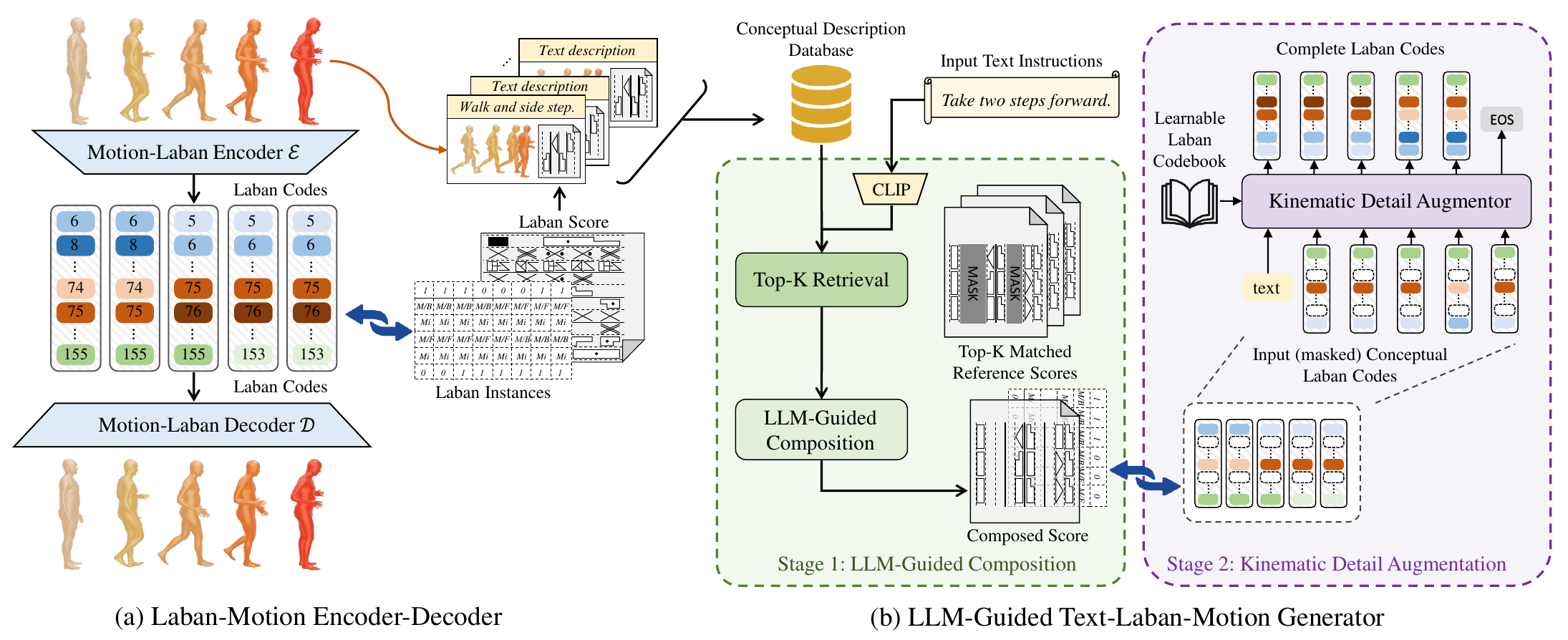}
\vspace{-3mm}
\caption{Overview of LaMoGen: (a) The Laban-motion Encoder-Decoder enables bidirectional conversion between motion and Laban instances. These instances are human-readable and LLM-editable, as each instance has a symbolic appearance and a conceptual description, stored in the Conceptual Description Database. (b) LLMs perform high-level symbolic planning via retrieval-augmented prompting, generating sequences of conceptual symbols. The Kinematic Detail Augmentor then enriches these sequences with details through autoregressive generation. Enriched symbol sequences are converted to instances, encoded as codes, and decoded into fine-grained motions.}
\vspace{-4mm}
\label{fig:framework}
\end{figure*}

\section{Methodology}
In this section, we first present LabanLite, an interpretable motion representation derived from Labanotation that encodes detailed body-part movements into symbolic form\footnote{Here, \textit{symbolic} refers to a discretized encoding system where each symbol from an annotation language carries specific semantic and domain-level information. This contrasts with continuous vector embeddings or with discrete representations, which are not inherently interpretable.}. Building on this, we introduce LaMoGen, a unified framework that leverages LabanLite to support LLM-driven motion generation. As shown in Fig.~\ref{fig:framework}, LaMoGen comprises two core components: a \textbf{Laban-Motion Encoder–Decoder Module} for bidirectional conversion between motions and Laban symbols, and a \textbf{LLM-Guided Text-Laban-Motion Generation Module}, where LLMs compose high-level symbolic motion sequences via retrieval-augmented prompting. These symbolic plans are then enriched with fine-grained symbols and attributes, which are subsequently decoded to produce motion, thus completing the text-driven motion generation process.

\vspace{-1mm}
\subsection{Core Concepts}\label{sec:core}
\vspace{-1mm}

\subsubsection{Labanotation and LabanLite} 
Labanotation is a symbolic system for recording human movement, where the staff is divided into eleven columns, each corresponding to a specific body part. Laban symbols are placed vertically on these columns to encode the movement type, affected body part, and motion duration~\citep{topaz1996elementary}. Symbols for less significant movements may be omitted, leaving blank spaces in the score (see score examples in Fig.~\ref{fig:teaser}). Since each symbol defines the onset and offset of a movement, this is known as event-wise annotation.

To enhance suitability for digital encoding and facilitate composition by LLMs in subsequent processes, we introduce LabanLite, which makes three enhancements over Labanotation while retaining its expressive capacity: 
\begin{enumerate}
\item Symbols are classified as either conceptual (representing primary movement structure) or detail (specifying fine-grained attributes)\footnote{E.g., the conceptual symbol for ``wave right hand'' specifies directional and level changes; the detailed symbol captures the arm's bend transitions.}; 
\item Event-wise annotation is replaced by frame-wise annotation, where each symbol is converted into a Laban instance at every frame, and staff columns are redefined as body-part groups with dedicated attribute fields\footnote{Because each body-part group has unique kinematic characteristics, their corresponding attribute fields are designed accordingly.}; 
\item Each conceptual symbol is accompanied by a strictly formatted textual description
(i.e., \texttt{<body-part group> <moving semantic> in <time> seconds})
termed \textit{\textbf{Conceptual Description}}. This standardized format enables unambiguous conversion from symbolic event-wise annotations to frame-wise instances through direct textual mapping.
\end{enumerate}

\noindent\textbf{Improved Temporal Expressiveness}. Unlike pose-based representations (e.g., CoMo~\citep{huang2024controllable}), LabanLite encodes both movement intention and duration, enabling language models to interpret and generate compositional instructions such as ``\textit{move hand forward in one second}''. These explicit and aligned symbolic attributes for direction and timing further allow LaMoGen to operate at the level of action units.

\noindent\textbf{Enhanced Interpretability}.
LabanLite utilizes a fixed set of symbols established by movement experts within Labanotation. For example, the \textit{Level} symbols comprise only three options (high, middle, low). Although finer levels could be introduced, we restrict the system to the existing symbols to ensure it remains both professionally relevant and interpretable by Laban experts.

\subsubsection{Laban code \& Laban codebook} 

After frame-wise annotation, each unique Laban instance for a specific body-part group is assigned a unique identifier, referred to as a \textbf{Laban code}. 
All $N$ codes are aggregated into the \textbf{Laban codebook} (\textit{detail explained in Supplementary Sec.~1}), denoted as ${C}=\{c_{n}\}_{n=1}^{N}$, where each entry $c_{n}\in\mathbb{R}^{d_c}$ is a learnable embedding.

The Laban codebook serves as a compact, trainable latent space capturing the fundamental semantics and kinematics of diverse Laban characteristics and attributes. Each Laban code thus acts as a symbolic anchor that can be mapped to a rich vectorized embedding, optimized during training to maximize reconstruction and compositional quality. By combining these embeddings linearly, the system can approximate continuous variations, enabling the composition of complex motions from simpler symbolic building blocks.

\subsection{Laban-Motion Encoder-Decoder}

Given a $T$-frame input motion: $X = \{x_t\}_{t=1}^{T}$, we propose an Automatic Symbol Detection Workflow $\mathcal{F}$, which converts $X$ into a Laban instance sequence $S=\mathcal{F}(X)$, and $S=\{s_{t}^{i,j}~|~t\in[1,T],~i\in[1,A_j]\,~j\in[1,G] \}$.
Here, $G$ is the total body-part group number, and $A_j$ is the attribute field dimension for Group $j$. 
The sequence $S$ is then encoded into a latent vector $Z$ using the Laban codebook ${C}$. The process will be explained as follows.

\subsubsection{Automatic Laban Code Detection Workflow}\label{sec:workflow}

The workflow processes each body-part group independently in three steps (\textit{see Supplementary Sec.~2 for better details}). On step one, \textit{\underline{Dynamic Interval Segmentation}}, we divide motion into coherent time intervals by classifying each frame as either dynamic or stationary (hold). This step ensures that symbolic units align with natural atomic actions.
On step two, \textit{\underline{Frame-wise Symbol Extraction}}, we translate motion signals into symbolic attributes. \textit{Direction} and \textit{Level} are measured by the 3D displacement of end-effectors from the pelvis, which are mapped to symbolic categories. \textit{Orientation} is obtained by calculating the Euler angles of the hip and discretising them into eight bins. \textit{Bend} is captured by discretising Euler angles between adjacent limbs into six bins, while \textit{Moving-effort} is quantified by discretising pelvis velocities into five levels. Together, these attributes provide a comprehensive symbolic account of local body dynamics.
On step three, in \textit{\underline{Interval-wise Symbol Aggregation}}, we assign a representative symbol combination to each segmented interval to ensure that the chosen symbol reflects the stable state of the motion.

\noindent\textbf{Standard Labanotation Discretization}. We implemented these rules directly according to conventionally accepted boundaries (e.g., orientation, bend angles, effort levels) in the Labanotation literature.

\subsubsection{Coding, Decoding \& Optimisation of Codebook}

The Laban-Motion Encoder $\mathcal{E}$ transforms the symbol sequence $S=\{s_t^{i,j}\}$, derived from the input motion $X=\{x_t\}$ as described in the previous section, into a latent representation:
    $z_{t} =\sum_{n=1}^{N}v_{t}^{n}c_{n},$
which is constructed by summing the embeddings of all active codes from the codebook $C=\{c_n\}_{n=1}^{N}$ at each frame.
The binary indicator vector $v_t \in \mathbb{R}^{N \times 1}$ specifies which codes are active: an entry $v_t^n$ is set to 1 whenever the detection workflow $\mathcal{F}$ identifies the corresponding attribute. 

We further implement a Laban-Motion Decoder $\mathcal{D}$ that reconstructs motion trajectories from the latent representations $\{z_t\}$, using a standard transformer decoder architecture~\citep{vaswani2017attention}. The Decoder parameters $\theta$ and the Laban codebook $C=\{c_n\}^N_{n=1}$ are jointly optimized by minimizing the following reconstruction loss $\mathcal{L}_{\text{rec}}$:
\begin{equation}
\{\theta,C\}= \underset{\theta,\{c_n\}^N_{n=1}}{\textrm{arg min }} \mathcal{L}_{rec}(X, \{\mathcal{D}(\sum_{n=1}^{N}v_{t}^{n}c_{n};\theta)\}_{t=1}^T), \\
\end{equation}
where $\mathcal{L}_{\text{rec}}$ is calculated by summing the $L_1$ distances between poses ($X$ and $\hat{X}$) and velocities ($\dot{X}$ and $\dot{\hat{X}}$) of the input and reconstructed motions:
$\mathcal{L}_{rec}(X,\hat{X}) = \|X - \hat{X}\|_1 + \lambda \|\dot{X} - \dot{\hat{X}}\|_1$,
with $\lambda$ being a hyperparameter that controls the relative weight of the two terms.

\subsection{LLM-Guided Text-Laban-Motion Generation}

Human motion can be understood as the outcome of both conceptual intent and physical execution. In LabanLite, \textit{conceptual symbols} capture high-level motion concepts and intentions (e.g., direction or level), whose structured patterns can be effectively modeled by large language models. \textit{Detail symbols}, by contrast, specify execution attributes such as orientation, bend, and effort—factors essential for realistic synthesis—which are modeled by a transformer-based architecture~\citep{zhang2023t2mgpt} via coding over the learned Laban Codebook. This hierarchical design naturally leads to a two-stage generation pipeline.

{
\setlength{\tabcolsep}{1mm}
\begin{table*}[t]
\centering\footnotesize
\caption{Quantitative comparisons on our Laban Benchmark, using the proposed Labanotation-based metrics, R-precision Top-3 (R@3) and FID. \textbf{Bold} and \underline{underlined} values indicate the best and the second-best performance, respectively.}
\vspace{-3mm}
\scalebox{0.9}{
\begin{tabular}{rrccccccccccccc}
\toprule
\multicolumn{2}{c}{\multirow{2}{*}{Method}} & \multicolumn{4}{c}{SMT $\uparrow$} & \multicolumn{4}{c}{TMP $\uparrow$} & \multicolumn{3}{c}{HMN $\uparrow$} & \multirow{2}{*}{R@3 $\uparrow$} & \multirow{2}{*}{FID $\downarrow$} \\
\cmidrule(lr){3-6} \cmidrule(lr){7-10} \cmidrule(lr){11-13}
& & supL & supR & armL & armR & supL & supR & armL & armR & arm-arm & arm-sup & sup-sup \\ 
\midrule
\multicolumn{2}{r}{Real data} & - & - & - & - & - & - & - & - & - & - & - & 0.216 & 0.001 \\
\cmidrule(lr){1-15}
\multicolumn{2}{r}{MDM~\citep{tevet2023mdm}} & 0.380 & 0.380 & 0.335 & 0.257 & 0.316 & 0.316 & 0.329 & 0.231 & 0.119 & 0.226 & 0.258 & 0.180 & 22.81 \\
\multicolumn{2}{r}{ReMoDiff~\citep{zhang2023remodiffuse}} & 0.470 & 0.470 & 0.427 & 0.395 & 0.377 & 0.377 & 0.385 & 0.322 & 0.179 & 0.264 & 0.351 & 0.192 & 7.121 \\
\multicolumn{2}{r}{MoDiff~\citep{zhang2024motiondiffuse}} & 0.491 & 0.491 & 0.470 & 0.411 & 0.362 & 0.362 & 0.412 & 0.328 & 0.180 & 0.281 & 0.361 & 0.196 & 5.701 \\
\multicolumn{2}{r}{CoMo~\citep{huang2024controllable}} & 0.358 & 0.358 & 0.474 & 0.382 & 0.211 & 0.211 & 0.284 & 0.250 & 0.203 & 0.298 & 0.252 & 0.176 & 21.94 \\
\multicolumn{2}{r}{MotionGPT~\citep{jiang2024motiongpt}} & 0.488 & 0.488 & 0.466 & 0.400 & 0.353 & 0.353 & 0.343 & 0.340 & 0.244 & 0.300 & 0.378 & 0.195 & 2.072 \\
\cmidrule(lr){1-15}
\multirow{5}{*}{\rotatebox{90}{\shortstack{LaMoGen\\Composer}}} & None & 0.523 & 0.523 & 0.430 & 0.392 & 0.337 & 0.337 & 0.361 & 0.385 & 0.215 & 0.356 & 0.393 & 0.199 & 5.562 \\
& Qwen3 & 0.571 & 0.571 & 0.478 & 0.495 & 0.401 & 0.401 & 0.456 & 0.448 & 0.369 & 0.334 & 0.450 & \textbf{0.212} & 1.903 \\
& DeepSeekV3 & 0.552 & 0.552 & \underline{0.496} & \underline{0.500} & 0.475 & 0.475 & 0.486 & 0.486 & \underline{0.370} & 0.326 & 0.463 & 0.206 & 1.859 \\
& GPT4.1 & \underline{0.583} & \underline{0.583} & {0.493} & 0.476 & \underline{0.507} & \underline{0.507} & \underline{0.501} & \underline{0.492} & {0.303} & \underline{0.367} & \underline{0.508} & \underline{0.208} & \underline{1.861} \\
& Human & \textbf{0.648} & \textbf{0.648} & \textbf{0.592} & \textbf{0.616} & \textbf{0.619} & \textbf{0.619} & \textbf{0.643} & \textbf{0.632} & \textbf{0.379} & \textbf{0.449} & \textbf{0.558} & {0.211} & \textbf{1.769} \\
\bottomrule
\vspace{-6mm}
\end{tabular}
}
\label{tab:loco}
\end{table*}
}

\begin{figure*}[t]
\centering
\includegraphics[width=0.93\linewidth]{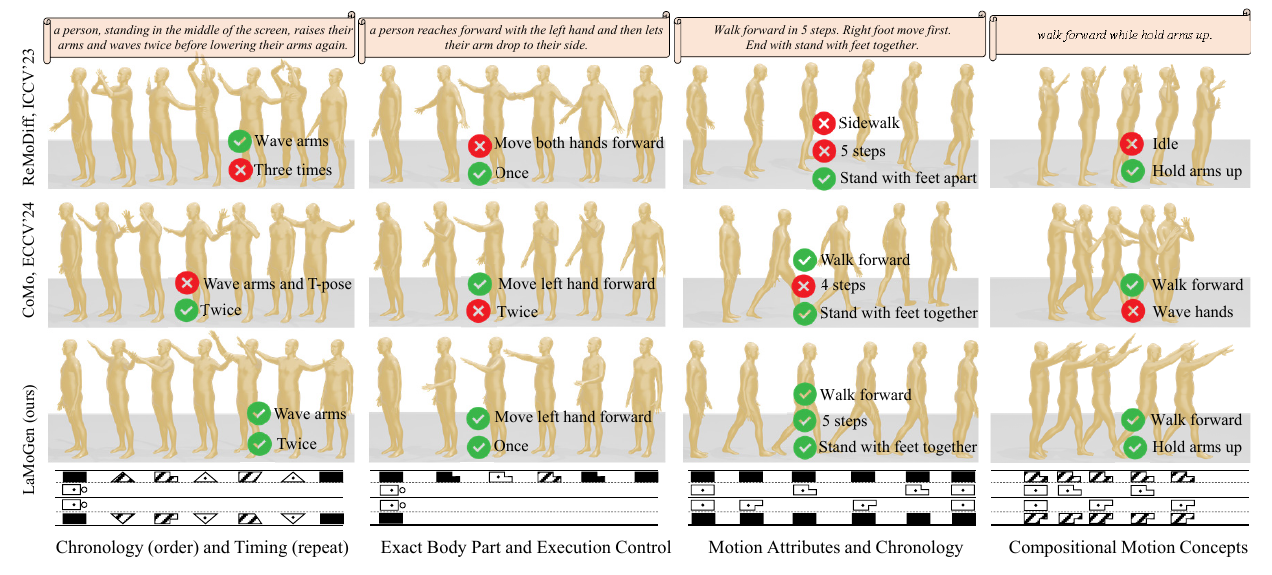}
\vspace{-4mm}
\caption{Qualitative comparisons on HumanML3D and HumanML3D-Laban test sets, with generated motions progressing from left to right. 
The LLM-composed conceptual symbols are shown below our results.
Misalignments between texts and motions are highlighted. Note how our method preserves correct sequencing, repetition, and timing of actions, while precisely controlling body parts and motion attributes, and demonstrating compositional generation enabled by LLM-driven symbolic inference—where existing methods fail.}
\vspace{-5mm}
\label{fig:qualitative}
\end{figure*}

{
\setlength{\tabcolsep}{1mm}
\begin{table*}[t]
\centering\footnotesize
\caption{Quantitative comparisons with state-of-the-art methods on the HumanML3D test and KIT-ML test datasets, under standard protocols. \textbf{Bold}, \underline{underlined}, and \textit{italicised} values denote the best, second-best, and third-best performance, respectively.}
\vspace{-3mm}
\scalebox{0.9}{
\begin{tabular}{rrrcccccccc}
\toprule
\multicolumn{3}{c}{\multirow{2}{*}{Method}} & \multicolumn{3}{c}{R-precision $\uparrow$} & \multirow{2}{*}{FID $\downarrow$} & \multirow{2}{*}{MM-Dist $\downarrow$} & \multirow{2}{*}{Diversity $\rightarrow$} & \multirow{2}{*}{Multi-Mod. $\uparrow$} \\ 
\cmidrule(lr){4-6}
& & & Top-1 & Top-2 & Top-3 \\ 
\midrule
\multirow{12}{*}{\rotatebox{90}{\underline{HumanML3D}}} & \multicolumn{2}{r}{Real data} & $0.511^{\pm .003}$ & $0.703^{\pm .003}$ & $0.797^{\pm .002}$ & $0.002^{\pm .000}$ & $2.974^{\pm .008}$ & $9.503^{\pm .085}$ & - & \\
\cmidrule(lr){2-11}
& \multicolumn{2}{r}{\citet{guo2022humanml3d}} & $0.457^{\pm .002}$ & $0.639^{\pm .003}$ & $0.740^{\pm .003}$ & $1.067^{\pm .002}$ & $3.340^{\pm .008}$ & $9.188^{\pm .002}$ & $\underline{2.090^{\pm .083}}$ & \\
& \multicolumn{2}{r}{MDM~\citep{tevet2023mdm}} & $0.320^{\pm .005}$ & $0.498^{\pm .004}$ & $0.611^{\pm .007}$ & $0.544^{\pm .044}$ & $5.566^{\pm .027}$ & $\mathbf{9.559^{\pm .086}}$ & $\mathbf{2.799^{\pm .072}}$ & \\
& \multicolumn{2}{r}{ReMoDiff~\citep{zhang2023remodiffuse}} & $\underline{0.510^{\pm .005}}$ & $\underline{0.698^{\pm .006}}$ & $\mathit{0.795^{\pm .004}}$ & $\mathbf{0.103^{\pm .004}}$ & $\mathbf{2.974^{\pm .016}}$ & $9.018^{\pm .075}$ & $1.795^{\pm .043}$ & \\
& \multicolumn{2}{r}{MoDiff~\citep{zhang2024motiondiffuse}} & $0.491^{\pm .001}$ & $0.681^{\pm .001}$ & $0.782^{\pm .001}$ & $0.630^{\pm .001}$ & $3.113^{\pm .001}$ & $\mathit{9.410^{\pm .049}}$ & $1.533^{\pm .042}$ & \\
& \multicolumn{2}{r}{CoMo~\citep{huang2024controllable}} & $\mathit{0.502^{\pm .002}}$ & $0.692^{\pm .007}$ & $0.790^{\pm .002}$ & $0.262^{\pm .004}$ & $\mathit{3.032^{\pm .015}}$ & $9.936^{\pm .066}$ & $1.013^{\pm .046}$ & \\
& \multicolumn{2}{r}{KP~\citep{liu2024bridging}} & $0.496^{\pm .000}$ & - & - & $0.275^{\pm .000}$ & - & $9.975^{\pm .000}$ & $\mathit{2.218^{\pm .000}}$ & \\
& \multicolumn{2}{r}{MotionGPT~\citep{jiang2024motiongpt}} & $0.492^{\pm .003}$ & $0.681^{\pm .003}$ & $0.778^{\pm .002}$ & $0.232^{\pm .008}$ & $3.096^{\pm .008}$ & $9.528^{\pm .071}$ & $2.008^{\pm .084}$ & \\
\cmidrule(lr){2-11}
& \multirow{4}{*}{\rotatebox{90}{\shortstack{LaMoGen\\Composer}}} & None & $0.438^{\pm .005}$ & $0.591^{\pm .004}$ & $0.755^{\pm .003}$ & $1.091^{\pm .013}$ & $3.999^{\pm .020}$ & $8.555^{\pm .097}$ & $1.421^{\pm .089}$ & \\
& & GPT4.1mini & $0.453^{\pm .003}$ & $0.679^{\pm .003}$ & $0.779^{\pm .004}$ & $0.561^{\pm .008}$ & $3.717^{\pm .011}$ & $8.434^{\pm .115}$ & $1.263 ^{\pm .034}$ & \\
& & GPT4.1 & $0.491^{\pm .002}$ & $\mathit{0.694^{\pm .002}}$ & $\underline{0.796^{\pm .003}}$ & $\mathit{0.252^{\pm .003}}$ & $3.087^{\pm .003}$ & $9.124^{\pm .058}$ & $1.131^{\pm .027}$ & \\
& & Human & $\mathbf{0.513^{\pm .003}}$ & $\mathbf{0.704^{\pm .002}}$ & $\mathbf{0.813^{\pm .006}}$ & $\underline{0.206^{\pm .003}}$ & $\underline{2.993^{\pm .009}}$ & $\underline{9.635^{\pm .109}}$ & $0.973^{\pm .024}$ & \\
\midrule
\multirow{12}{*}{\rotatebox{90}{{\underline{KIT-ML}}}} & \multicolumn{2}{r}{Real data} & $0.424^{\pm .005}$ & $0.649^{\pm .006}$ & $0.779^{\pm .006}$ & $0.031^{\pm .006}$ & $2.788^{\pm .012}$ & $11.08^{\pm .097}$ & - & \\
\cmidrule(lr){2-11}
& \multicolumn{2}{r}{Guo et al.~\citep{guo2022humanml3d}} & $0.370^{\pm .005}$ & $0.569^{\pm .007}$ & $0.693^{\pm .007}$ & $2.770^{\pm .109}$ & $3.401^{\pm .008}$ & $10.91^{\pm .119}$ & $\underline{1.482^{\pm .065}}$ & \\
& \multicolumn{2}{r}{MDM~\citep{tevet2023mdm}} & $0.164^{\pm .004}$ & $0.291^{\pm .004}$ & $0.396^{\pm .004}$ & $0.497^{\pm .021}$ & $9.191^{\pm .022}$ & $10.85^{\pm .109}$ & $\mathbf{1.907^{\pm .214}}$ & \\
& \multicolumn{2}{r}{ReMoDiff~\citep{zhang2023remodiffuse}} & $\mathbf{0.427^{\pm .014}}$ & $\mathit{0.641^{\pm .004}}$ & $0.765^{\pm .055}$ & $\mathbf{0.155^{\pm .006}}$ & $\mathbf{2.814^{\pm .012}}$ & $10.80^{\pm .105}$ & $1.239^{\pm .028}$ & \\
& \multicolumn{2}{r}{MoDiff~\citep{zhang2024motiondiffuse}} & $0.417^{\pm .004}$ & $0.621^{\pm .004}$ & $0.739^{\pm .004}$ & $1.954^{\pm .062}$ & $2.958^{\pm .005}$ & \underline{${11.10^{\pm .143}}$} & $0.730^{\pm .013}$ & \\
& \multicolumn{2}{r}{CoMo~\citep{huang2024controllable}} & $\mathit{0.422^{\pm .009}}$ & $0.638^{\pm .007}$ & $\mathit{0.765^{\pm .011}}$ & $\mathit{0.332^{\pm .045}}$ & $\mathit{2.873^{\pm .021}}$ & $\mathit{10.95^{\pm .196}}$ & $1.249^{\pm .008}$ & \\
& \multicolumn{2}{r}{MotionGPT~\citep{jiang2024motiongpt}} & $0.366^{\pm .005}$ & $0.558^{\pm .004}$ & $0.680^{\pm .005}$ & $0.510^{\pm .008}$ & $3.527^{\pm .021}$ & $10.35^{\pm .084}$ & $2.328^{\pm .117}$ & \\
\cmidrule(lr){2-11}
& \multirow{4}{*}{\rotatebox{90}{\shortstack{LaMoGen\\Composer}}} & None & $0.400^{\pm .004}$ & $0.621^{\pm .005}$ & $0.750^{\pm .003}$ & $0.685^{\pm .005}$ & $3.222^{\pm .011}$ & $11.74^{\pm .140}$ & $\mathit{1.305^{\pm .122}}$ & \\
& & GPT4.1mini & $0.418^{\pm .005}$ & $0.630^{\pm .004}$ & ${0.761}^{\pm .006}$ & $0.550^{\pm .005}$ & $3.274^{\pm .011}$ & $11.85^{\pm .175}$ & $1.103^{\pm .051}$ & \\
& & GPT4.1 & $0.421^{\pm .004}$ & $\underline{0.649^{\pm .004}}$ & $\underline{0.775^{\pm .013}}$ & $0.415^{\pm .011}$ & $3.165^{\pm .007}$ & $11.30^{\pm .166}$ & $1.028^{\pm .101}$ & \\
& & Human & $\underline{0.424^{\pm .006}}$ & $\mathbf{0.657^{\pm .005}}$ & $\mathbf{0.782^{\pm .009}}$ & $\underline{0.254^{\pm .004}}$ & $\underline{2.821^{\pm .094}}$ & $\mathbf{11.09^{\pm .184}}$ & $0.979^{\pm .012}$ & \\
\bottomrule
\vspace{-5mm}
\end{tabular}
}
\label{tab:t2m}
\end{table*}
}

\subsubsection{LLM-Guided Motion Concept Composition}

As introduced earlier, Laban symbols are both visually interpretable and structurally organized, making them well-suited for symbolic reasoning with language models. 
Leveraging these properties, the first stage of our generation framework employs LLMs to compose motion plans at the conceptual level via retrieval-augmented prompting. 

Specifically, as depicted in Fig.~\ref{fig:framework}(b), we maintain a Conceptual Description Database where each record consists of a motion caption (the primary key) and its corresponding conceptual description (the value). Motion captions may be descriptive or prescriptive, accommodating both vague and explicit user intents.
During inference, a user-provided query is processed by computing its semantic similarity to the database captions using CLIP~\cite{radford2021clip}. The top-$K$ retrieved entries, which are expected to cover the relevant movement elements expressed by the user, are provided to the LLM as in-context examples along with the incoming query. The LLM then reasons over the alignment between textual instructions and symbolic motion patterns, editing or composing a new conceptual description tailored to the user's input. Owing to the standardized format of conceptual descriptions, these can be unambiguously mapped back to Laban symbols, enabling effective symbolic motion synthesis from textual input.

\subsubsection{Kinematic Detail Augmentation over Codebook}

While LLMs are effective at high-level planning, they lack the temporal modelling precision required for detailed motion synthesis. Thus, in the second stage, the Kinematic Detail Augmentor autoregressively enriches conceptual symbol sequences with detail symbols, expanding them into \textit{temporally} coherent and \textit{complete} LabanLite codes. This process adds around 60\% more information to the symbol sequence, supporting temporally coherent and detailed kinematic motion.

Specifically, we represent the conceptual symbol sequence from the first stage as masked binary indicator vectors $\hat{v}_{1:t-1}$, where only the predicted conceptual symbol fields are activated. Conditioned on the text input $m$ and $\hat{v}_{1:t-1}$, the Augmentor predicts full binary indicator vectors $v_t$ for each frame, activating embedding entries in the Laban codebook, which encodes both conceptual and detail attributes. The prediction is made over estimated activation probabilities: $p_t^n = P(v_t^n = 1 | m, \hat{v}_{1:t-1})$. 

During training, conceptual vectors are constructed by masking detailed attributes in the input, with random masking applied to improve generalisation and prevent overfitting. To mark motion termination, an end-of-sequence token $\texttt{<EOS>}$ is appended, extending the codebook to $N+1$ entries. Through this process, the Augmentor enriches the frame-wise conceptual symbol plans produced by the LLM with detailed attributes, converting them into fully specified Laban codes. The learning objective is defined using a binary cross-entropy loss:
\begin{equation}
\mathcal{L}_{gen} = -\sum_{t,n} \left[ v_t^n \log p_t^n + (1-v_t^n) \log(1-p_t^n)  \right].
\end{equation}
Finally, the Laban-Motion Decoder $\mathcal{D}$ reconstructs the enriched codes into instruction-aligned, detailed motion trajectories.

\section{Laban Benchmark}

To evaluate the effectiveness of the proposed LaMoGen framework—specifically its capacity to capture fine-grained, structured textual instructions and accurately reflect them in generated motion—we introduce a new benchmark. This benchmark comprises a motion–Laban–text paired dataset, along with three Labanotation-based metrics designed to assess symbolic, temporal, and harmonious alignment. \textit{Further details regarding dataset construction and evaluation metrics are provided in Supplementary Sec.~3.}

\noindent\textbf{Motion–Laban–Text Paired Dataset.}
This dataset consists of triplets, each containing (1) a motion sequence, (2) its corresponding Laban symbol sequence, and (3) a textual instructional description. The motion data is derived from HumanML3D~\cite{guo2022humanml3d}, forming the \textit{HumanML3D-Laban} corpus. We selected motions amenable to instructional description, resulting in a predominance of locomotion sequences, which are further decomposed into atomic actions. Annotation is semi-automated: motion details such as step count, left/right step order, and action labels are manually extracted and then expanded into natural language descriptions with the help of LLMs. 

\noindent\textbf{Laban-based Metrics.}
Given the professional rigor of Labanotation in annotating and analyzing motion, we propose three Laban metrics, i.e., Semantic Alignment (SMT), Temporal Alignment (TMP), and Harmonious Alignment (HMN), to assess the accuracy of orientation, duration, and coordination in fine-grained text-driven motion generation. Unlike KP benchmark~\citep{liu2024bridging}, which is limited to one or two body parts, our metrics evaluate multi-body consistency across four key body parts. These metrics operate by converting both ground-truth and generated motions into conceptual symbol sequences, followed by intra- and inter-body-part comparisons using \textit{Longest Common Subsequence (See mathematical definition in Supplementary Sec.~3)}.
Specifically,
SMT measures sequence similarity within individual body parts while ignoring duration. TMP extends this by incorporating symbol duration to enforce temporal consistency. HMN further evaluates coordination across multiple body parts by treating co-occurring symbols as joint units. For example, if the ground truth shows the left foot stepping forward while the left arm swings backward, the generated motion is expected to exhibit a comparable synchronised pattern.

\section{Experiments}

\subsection{Experimental Setup}

We assess the automatic Laban detection workflow using a Laban-annotated motion dataset~\citep{jiang2024motion}. This dataset is manually annotated by Laban experts and consists of 21 walking motions. As a baseline, we employ the hardcoded detection algorithm from \citet{ikeuchi2018describing}, extending its original upper-body focus to full-body motions by applying the same thresholds. Detection accuracy is evaluated using the three proposed Laban metrics.

For motion generation, following standard protocols~\citep{guo2022humanml3d}, we assess LaMoGen on HumanML3D~\citep{guo2022humanml3d} and KIT-ML~\citep{plappert2016kit} datasets.
We compare our approach against state-of-the-art methods across different paradigms, including: (1) two baseline methods, \citet{guo2022humanml3d} and MDM~\citep{tevet2023mdm}; (2) a latent retrieval-augmented method,  ReMoDiff~\citep{zhang2023remodiffuse}; (3) a fine-grained generation method, MoDiff~\citep{zhang2024motiondiffuse};  (4) two explicit motion decomposition methods, KP~\citep{liu2024bridging} and CoMo~\cite{huang2024controllable}; and (5) an LLM-finetuned method, MotionGPT~\citep{jiang2024motiongpt}.
Evaluation metrics include: Fr\'echet Inception Distance (FID) for distributional similarity (lower is better); R-Precision and Multimodal Distance (MM-Dist) for text-motion correspondence (higher is better); and Diversity and Multi-Modality (Multi-Mod.) for motion variability (Diversity closer to ground truth and higher Multi-Mod. are preferred). For the Laban Benchmark, we evaluate models trained on HumanML3D without fine-tuning, reporting Laban metrics, R-Precision, and FID for comprehensive assessment. All results are averaged over 20 independent runs.

{
\setlength{\tabcolsep}{0.5mm}
\begin{table}[t]
\centering
\caption{Performance of Laban symbol detection on the Laban-annotated motion dataset, using average Laban metrics. \textbf{Bold} values denote the best performance.}
\vspace{-3mm}
\scalebox{0.9}{
\begin{tabular}{cccc}
\toprule
Method & avg. SMT $\uparrow$ & avg. TMP $\uparrow$ & avg. HMN $\uparrow$ \\
\midrule
\citet{ikeuchi2018describing} & 0.751 & 0.632 & 0.611 \\
Ours & \textbf{0.871} & \textbf{0.852} & \textbf{0.786} \\
\bottomrule
\vspace{-10mm}
\end{tabular}}
\label{tab:laban_acc}
\end{table}
}

\subsection{Evaluation on Laban Detection}

Table~\ref{tab:laban_acc} reports the Laban symbol detection accuracy, demonstrating that our detection workflow achieves high accuracy. This enables robust and precise conversion of motion data into Laban symbol sequences, providing high-quality data for the Conceptual Description Database and the construction of the Laban Benchmark.

\subsection{Evaluation on Motion Generation}

Tables~\ref{tab:loco}, \ref{tab:t2m} report quantitative comparisons between our method and state-of-the-art methods on the HumanML3D-Laban, HumanML3D, and KIT-ML datasets. We evaluate our model under six high-level composer conditions: (1) \textit{None}, where motion generation is conditioned solely on text and serves as a baseline for text comprehension; (2) three LLM composers (\textit{GPT4.1mini}, \textit{GPT4.1}, \textit{Qwen3}, \textit{DeepSeekV3}), where generation is conditioned on both text and conceptual cues provided by LLMs; and (3) \textit{Human}, in which conceptual cues are derived from ground-truth annotations to simulate human composition.

\noindent\textbf{Results on Laban Benchmark.}
As shown in Table~\ref{tab:loco}, LaMoGen achieves the best performance across both Laban metrics and two standard evaluation metrics, demonstrating superior text-motion alignment compared to other methods. 
The LLM-assisted configurations show that LLMs can effectively interpret and compose conceptual Laban symbols as explicit guidance, resulting in higher Laban metric scores than other baselines. In contrast, methods such as MDM and CoMo, that rely on joint text–motion embeddings, often fail to align text and motion under lengthy or unseen inputs, as reflected by their higher FID scores.

Furthermore, as evidenced in Table~\ref{tab:loco} and \ref{tab:t2m}, MotionGPT, while less competitive on conventional benchmarks, exhibits a stronger ability to understand instructional descriptions on the Laban Benchmark—surpassing CoMo in this aspect. This finding supports the argument presented in KP~\citep{liu2024bridging}, namely that conventional metrics fail to accurately distinguish the motion generation capabilities of existing methods. It further underscores the necessity of the proposed Laban metrics for a more comprehensive evaluation of generative motion quality.

\noindent\textbf{Results on conventional benchmarks.}
As shown in Table~\ref{tab:t2m}, LaMoGen consistently ranks top-3 across five standard metrics, highlighting its robust generalisation.
We attribute these strong results to the proposed LabanLite representation, which discretises motion into interpretable symbolic sequences. This design enhances the consistency between text and motion, and enables the model to reason about both inter- and intra-body part relationships. 

However, although LaMoGen excels on most metrics, its FID scores lag behind others. We believe the reason is that its high abstraction uses identical symbols for different low-level variations that share the same high-level semantics. For example, raising the hand from ``Lo.'' to ``Hi.'' is represented by the same symbols, but different individuals may perform it with varying speeds. 
Such low-level movement variations are inherently beyond LabanLite's expressive capacity, resulting in higher FID scores.

\noindent\textbf{Qualitative comparisons.}
We evaluate the motion in four aspects: \textit{Chronology}, whether motions occur in the proper sequence, such as which body part moves first; \textit{Timing control}, whether motions follow specified durations; \textit{Explainability}, whether it is clear why motions happen a certain way; and \textit{Compositionality}, whether different body-part actions can be combined. 
Four examples are shown in Fig.~\ref{fig:qualitative}, \textit{with more comparisons in Supplementary Sec.~5}.

Our analysis reveals that ReMoDiff and CoMo can violate chronological order and struggle with precise movement timing. CoMo ends in a T-pose in the first example, and ReMoDiff moves the hands three times in the second. Without explicit control signals, their motions are less explainable—it is unclear why ReMoDiff moves both hands forward. In contrast, LoMoGen uses symbolic representations to clarify movement logic, such as when to step forward and which body part acts. Regarding compositionality, in the final example, ReMoDiff raises the arms but does not walk, CoMo walks while waving the arms, whereas our method combines both actions correctly. Across all examples, our method satisfies all four criteria, highlighting the advantage of LLM-driven symbolic inference.

We further examine each method's ability to model fine-grained temporal structure by modifying input text with either vague (e.g., speed change) or precise (e.g., time duration) descriptions. Results highlight our method excels, as LLMs adapt conceptual descriptions while others lack this compositional capability.

\noindent\textbf{User studies.} We conduct a user study to evaluate the generated motion quality of LaMoGen against ReMoDiff and CoMo. Results show that participants consistently preferred LaMoGen over other methods.

\vspace{-1mm}
\subsection{Ablation study}
\vspace{-1mm}

\noindent\textbf{LLM's capability}.
We evaluate two open-source LLMs and two versions of GPT-4.1 to assess how model capability influences performance. Table~\ref{tab:loco} and \ref{tab:t2m} reveal that stronger LLMs achieve better generation, more effectively composing Laban symbols and understanding their relationships.

\noindent\textbf{Number of top-matched references}.
To guide LLMs in composing conceptual codes, we provide top-matched examples and investigate the impact of example quantity on HumanML3D test set performance using GPT-4.1. 
Results (\textit{in Supplementary Sec.~5}) show that increasing examples from 1 to 3 consistently improves performance, indicating that LLMs benefit from having sufficient examples for accurate imitation. But adding more (5 or 7) offers no further improvement, likely due to exceeding the LLM’s context window and causing it to forget the most relevant cues. Thus, we use top-3 retrieval as the default setting.

\noindent\textbf{Masking ratio on Laban codes}.
We further analyze how varying the masking ratio on Laban codes influences generation, using GPT-4.1 and top-3 retrieval on HumanML3D. A higher masking ratio reduces the influence of conceptual cues, making motion generation rely more on text and less on Laban symbols, reducing the influence of conceptual guidance. Results indicate that a masking ratio of 0.3 offers the best balance and achieves optimal performance.

\vspace{-1mm}
\section{Conclusion}
\vspace{-1mm}

In this work, we have introduced LabanLite, a novel, human-interpretable motion representation rooted in Labanotation, and presented \textbf{LaMoGen}, a pioneering Text-to-\textbf{La}banotation-to-\textbf{Mo}tion \textbf{Gen}eration framework. By leveraging LabanLite’s concise symbolic abstraction, our approach enables large language models to autonomously plan and compose motion sequences through explicit, interpretable symbolic reasoning—moving beyond the limitations of traditional joint embedding methods. The newly proposed Labanotation-based benchmark, together with comprehensive metrics, provides a rigorous, multidimensional evaluation of text-motion alignment in symbolic, temporal, and harmony aspects.
Experiments on both our benchmark and public datasets demonstrate that LaMoGen achieves state-of-the-art performance in terms of interpretability and controllability, effectively capturing fine-grained temporal structures and explicit action sequences from natural language instructions. 

\section*{Acknowledgments}
This research was supported by the Theme-based Research Scheme, Research Grants Council of Hong Kong (T45-205/21-N), and the Guangdong and Hong Kong Universities “1+1+1” Joint Research Collaboration Scheme (2025A0505000003). The authors would like to thank Ms. Wendy Chu Mang-Ching from the School of Dance, The Hong Kong Academy for Performing Arts for the training and discussions on Labanotation.

{
    \small
    \bibliographystyle{ieeenat_fullname}
    \bibliography{main}

@String{Computer = "{IEEE} Computer" }

@String{Springer = "Springer-Verlag" }

@inproceedings{athanasiou2022teach,
  title={Teach: Temporal action composition for 3d humans},
  author={Athanasiou, Nikos and Petrovich, Mathis and Black, Michael J and Varol, G{\"u}l},
  booktitle={3DV},
  pages={414--423},
  year={2022},
  organization={IEEE}
}

@article{jiang2024motiongpt,
  title={Motiongpt: Human motion as a foreign language},
  author={Jiang, Biao and Chen, Xin and Liu, Wen and Yu, Jingyi and Yu, Gang and Chen, Tao},
  journal={Advances in Neural Information Processing Systems},
  volume={36},
  year={2024}
}

@inproceedings{zhang2023remodiffuse,
  title={Remodiffuse: Retrieval-augmented motion diffusion model},
  author={Zhang, Mingyuan and Guo, Xinying and Pan, Liang and Cai, Zhongang and Hong, Fangzhou and Li, Huirong and Yang, Lei and Liu, Ziwei},
  booktitle={ICCV},
  pages={364--373},
  year={2023}
}

@inproceedings{karunratanakul2023gmd,
  title={Guided motion diffusion for controllable human motion synthesis},
  author={Karunratanakul, Korrawe and Preechakul, Konpat and Suwajanakorn, Supasorn and Tang, Siyu},
  booktitle={ICCV},
  pages={2151--2162},
  year={2023}
}

@article{plappert2016kit,
  title={The kit motion-language dataset},
  author={Plappert, Matthias and Mandery, Christian and Asfour, Tamim},
  journal={Big data},
  volume={4},
  number={4},
  pages={236--252},
  year={2016},
  publisher={Mary Ann Liebert, Inc. 140 Huguenot Street, 3rd Floor New Rochelle, NY 10801 USA}
}

@article{ikeuchi2018describing,
  title={Describing upper-body motions based on labanotation for learning-from-observation robots},
  author={Ikeuchi, Katsushi and Ma, Zhaoyuan and Yan, Zengqiang and Kudoh, Shunsuke and Nakamura, Minako},
  journal={International Journal of Computer Vision},
  volume={126},
  pages={1415--1429},
  year={2018}
}

@book{topaz1996elementary,
  title={Elementary Labanotation: A study guide},
  author={Topaz, Muriel and Blum, Odette and Segall, Jessica},
  publisher={Dance Notation Bureau},
  year={1996}
}

@inproceedings{SMPL-X:2019,
  title = {Expressive Body Capture: {3D} Hands, Face, and Body from a Single Image},
  author = {Pavlakos, Georgios and Choutas, Vasileios and Ghorbani, Nima and Bolkart, Timo and Osman, Ahmed A. A. and Tzionas, Dimitrios and Black, Michael J.},
  booktitle = {Proceedings of the IEEE/CVF conference on computer Vision and Pattern Recognition},
  pages     = {10975--10985},
  year = {2019}
}

@inproceedings{jiang2024motion,
  title={Motion Part-Level Interpolation and Manipulation over Automatic Symbolic Labanotation Annotation},
  author={Jiang, Junkun and Au, Ho Yin and Chen, Jie and Xiang, Jingyu and Chen, Mingyuan},
  booktitle={IJCNN},
  pages={1--8},
  year={2024},
  organization={IEEE}
}

@inproceedings{
tevet2023mdm,
title={Human Motion Diffusion Model},
author={Guy Tevet and Sigal Raab and Brian Gordon and Yoni Shafir and Daniel Cohen-or and Amit Haim Bermano},
booktitle={The Eleventh International Conference on Learning Representations },
year={2023},
url={https://openreview.net/forum?id=SJ1kSyO2jwu}
}

@inproceedings{guo2022humanml3d,
  title={Generating diverse and natural 3d human motions from text},
  author={Guo, Chuan and Zou, Shihao and Zuo, Xinxin and Wang, Sen and Ji, Wei and Li, Xingyu and Cheng, Li},
  booktitle={CVPR},
  pages={5152--5161},
  year={2022}
}

@article{zhang2024motiondiffuse,
  title={Motiondiffuse: Text-driven human motion generation with diffusion model},
  author={Zhang, Mingyuan and Cai, Zhongang and Pan, Liang and Hong, Fangzhou and Guo, Xinying and Yang, Lei and Liu, Ziwei},
  journal={IEEE Transactions on Pattern Analysis and Machine Intelligence},
  year={2024},
  publisher={IEEE}
}

@inproceedings{radford2021clip,
  title={Learning transferable visual models from natural language supervision},
  author={Radford, Alec and Kim, Jong Wook and Hallacy, Chris and Ramesh, Aditya and Goh, Gabriel and Agarwal, Sandhini and Sastry, Girish and Askell, Amanda and Mishkin, Pamela and Clark, Jack and others},
  booktitle={International conference on machine learning},
  pages={8748--8763},
  year={2021},
  organization={PMLR}
}

@inproceedings{huang2024controllable,
  title={Como: Controllable motion generation through language guided pose code editing},
  author={Huang, Yiming and Wan, Weilin and Yang, Yue and Callison-Burch, Chris and Yatskar, Mark and Liu, Lingjie},
  booktitle={ECCV},
  pages={180--196},
  year={2024},
  organization={Springer}
}

@inproceedings{liu2024bridging,
  title={Bridging the gap between human motion and action semantics via kinematic phrases},
  author={Liu, Xinpeng and Li, Yong-Lu and Zeng, Ailing and Zhou, Zizheng and You, Yang and Lu, Cewu},
  booktitle={ECCV},
  pages={223--240},
  year={2024},
  organization={Springer}
}

@article{li2023labanformer,
  title={LabanFormer: Multi-scale graph attention network and transformer with gated recurrent positional encoding for labanotation generation},
  author={Li, Min and Miao, Zhenjiang and Lu, Yuanyao},
  journal={Neurocomputing},
  volume={539},
  pages={126203},
  year={2023},
  publisher={Elsevier}
}

@inproceedings{li2024translating,
  title={Translating Motion to Notation: Hand Labanotation for Intuitive and Comprehensive Hand Movement Documentation},
  author={Li, Ling and Yang, WenRui and Yu, Xinchun and Xing, Junliang and Zhang, Xiao-Ping},
  booktitle={Proceedings of the 32nd ACM International Conference on Multimedia},
  pages={4092--4100},
  year={2024}
}

@article{delmas2024posescript,
  title={Posescript: Linking 3d human poses and natural language},
  author={Delmas, Ginger and Weinzaepfel, Philippe and Lucas, Thomas and Moreno-Noguer, Francesc and Rogez, Gr{\'e}gory},
  journal={IEEE transactions on pattern analysis and machine intelligence},
  year={2024},
  publisher={IEEE}
}

@inproceedings{zhang2023t2mgpt,
  title={Generating human motion from textual descriptions with discrete representations},
  author={Zhang, Jianrong and Zhang, Yangsong and Cun, Xiaodong and Zhang, Yong and Zhao, Hongwei and Lu, Hongtao and Shen, Xi and Shan, Ying},
  booktitle={CVPR},
  pages={14730--14740},
  year={2023}
}

@inproceedings{wan2024tlcontrol,
  title={Tlcontrol: Trajectory and language control for human motion synthesis},
  author={Wan, Weilin and Dou, Zhiyang and Komura, Taku and Wang, Wenping and Jayaraman, Dinesh and Liu, Lingjie},
  booktitle={ECCV},
  pages={37--54},
  year={2024},
  organization={Springer}
}

@article{liu2023plan,
  title={Plan, posture and go: Towards open-world text-to-motion generation},
  author={Liu, Jinpeng and Dai, Wenxun and Wang, Chunyu and Cheng, Yiji and Tang, Yansong and Tong, Xin},
  journal={arXiv preprint arXiv:2312.14828},
  year={2023}
}

@inproceedings{wang2025stickmotion,
  title={StickMotion: Generating 3D Human Motions by Drawing a Stickman},
  author={Wang, Tao and Wu, Zhihua and He, Qiaozhi and Chu, Jiaming and Qian, Ling and Cheng, Yu and Xing, Junliang and Zhao, Jian and Jin, Lei},
  booktitle={CVPR},
  pages={12370--12379},
  year={2025}
}

@article{shi2023generating,
  title={Generating fine-grained human motions using chatgpt-refined descriptions},
  author={Shi, Xu and Luo, Chuanchen and Peng, Junran and Zhang, Hongwen and Sun, Yunlian},
  journal={arXiv preprint arXiv:2312.02772},
  volume={1},
  year={2023}
}

@inproceedings{athanasiou2023sinc,
  title={SINC: Spatial composition of 3D human motions for simultaneous action generation},
  author={Athanasiou, Nikos and Petrovich, Mathis and Black, Michael J and Varol, G{\"u}l},
  booktitle={CVPR},
  pages={9984--9995},
  year={2023}
}

@article{achiam2023gpt,
  title={Gpt-4 technical report},
  author={Achiam, Josh and Adler, Steven and Agarwal, Sandhini and Ahmad, Lama and Akkaya, Ilge and Aleman, Florencia Leoni and Almeida, Diogo and Altenschmidt, Janko and Altman, Sam and Anadkat, Shyamal and others},
  journal={arXiv preprint arXiv:2303.08774},
  year={2023}
}

@article{vaswani2017attention,
  title={Attention is all you need},
  author={Vaswani, Ashish and Shazeer, Noam and Parmar, Niki and Uszkoreit, Jakob and Jones, Llion and Gomez, Aidan N and Kaiser, {\L}ukasz and Polosukhin, Illia},
  journal={Advances in neural information processing systems},
  volume={30},
  year={2017}
}

@inproceedings{zhang2024motiongpt,
  title={Motiongpt: Finetuned llms are general-purpose motion generators},
  author={Zhang, Yaqi and Huang, Di and Liu, Bin and Tang, Shixiang and Lu, Yan and Chen, Lu and Bai, Lei and Chu, Qi and Yu, Nenghai and Ouyang, Wanli},
  booktitle={AAAI},
  volume={38},
  number={7},
  pages={7368--7376},
  year={2024}
}

@inproceedings{zhou2024avatargpt,
  title={Avatargpt: All-in-one framework for motion understanding planning generation and beyond},
  author={Zhou, Zixiang and Wan, Yu and Wang, Baoyuan},
  booktitle={CVPR},
  pages={1357--1366},
  year={2024}
}

@inproceedings{zeng2025light,
  title={Light-t2m: A lightweight and fast model for text-to-motion generation},
  author={Zeng, Ling-An and Huang, Guohong and Wu, Gaojie and Zheng, Wei-Shi},
  booktitle={AAAI},
  volume={39},
  number={9},
  pages={9797--9805},
  year={2025}
}

@inproceedings{lee2023multiact,
  title={Multiact: Long-term 3d human motion generation from multiple action labels},
  author={Lee, Taeryung and Moon, Gyeongsik and Lee, Kyoung Mu},
  booktitle={AAAI},
  volume={37},
  number={1},
  pages={1231--1239},
  year={2023}
}

@inproceedings{kalakonda2023action,
  title={Action-gpt: Leveraging large-scale language models for improved and generalized action generation},
  author={Kalakonda, Sai Shashank and Maheshwari, Shubh and Sarvadevabhatla, Ravi Kiran},
  booktitle={ICME},
  pages={31--36},
  year={2023},
  organization={IEEE}
}

@misc{wordcloud,
  title={wordcloud: A little word cloud generator in Python},
  author = {Andreas Mueller},
  year={2015},
  howpublished = {\url{https://github.com/amueller/word_cloud}},
  note={Accessed: 2025-06-30}
}

@article{li2024lamp,
  title={Lamp: Language-motion pretraining for motion generation, retrieval, and captioning},
  author={Li, Zhe and Yuan, Weihao and He, Yisheng and Qiu, Lingteng and Zhu, Shenhao and Gu, Xiaodong and Shen, Weichao and Dong, Yuan and Dong, Zilong and Yang, Laurence T},
  journal={arXiv preprint arXiv:2410.07093},
  year={2024}
}

@article{qwen3,
  title={Qwen3 technical report},
  author={Yang, An and Li, Anfeng and Yang, Baosong and Zhang, Beichen and Hui, Binyuan and Zheng, Bo and Yu, Bowen and Gao, Chang and Huang, Chengen and Lv, Chenxu and others},
  journal={arXiv preprint arXiv:2505.09388},
  year={2025}
}

@misc{deepseekv3,
      title={DeepSeek-V3 Technical Report}, 
      author={DeepSeek-AI},
      year={2024},
      eprint={2412.19437},
      archivePrefix={arXiv},
      primaryClass={cs.CL},
      url={https://arxiv.org/abs/2412.19437}, 
}

@inproceedings{guo2024momask,
  title={Momask: Generative masked modeling of 3d human motions},
  author={Guo, Chuan and Mu, Yuxuan and Javed, Muhammad Gohar and Wang, Sen and Cheng, Li},
  booktitle={Proceedings of the IEEE/CVF Conference on Computer Vision and Pattern Recognition},
  pages={1900--1910},
  year={2024}
}

@article{au2025soscontrol,
  title={SOSControl: Enhancing Human Motion Generation through Saliency-Aware Symbolic Orientation and Timing Control},
  author={Au, Ho Yin and Jiang, Junkun and Chen, Jie},
  journal={arXiv preprint arXiv:2601.14258},
  year={2025}
}

@article{au2025deep,
  title={Deep Compositional Phase Diffusion for Long Motion Sequence Generation},
  author={Au, Ho Yin and Chen, Jie and Jiang, Junkun and Xiang, Jingyu},
  journal={arXiv preprint arXiv:2510.14427},
  year={2025}
}

@article{jiang2026mmdm,
      title={Learning Context-Adaptive Motion Priors for Masked Motion Diffusion Models with Efficient Kinematic Attention Aggregation}, 
      author={Junkun Jiang and Jie Chen and Ho Yin Au and Jingyu Xiang},
      year={2026},
      journal={https://arxiv.org/abs/2603.07697}, 
}
}


\clearpage

\twocolumn[
    \begin{@twocolumnfalse}
        \customtitle
    \end{@twocolumnfalse}
]
\setcounter{section}{0}
\setcounter{figure}{0}
\setcounter{table}{0}

\footnotetext[1]{%
  \begin{minipage}[t]{\columnwidth}
  Corresponding author. \\
  Project page: \url{https://jjkislele.github.io/LaMoGen/}
  \end{minipage}
}

This supplementary material provides additional details that could not be included in the main paper due to page limitations. The contents are organized as follows:
\begin{itemize}
\item \textbf{Details of LabanLite} (Sec.~\ref{sec:sup_labanlite}): Includes a comprehensive definition of LabanLite, semantic interpretations of Laban symbols, as well as the relationships among Laban symbols and Laban instances. Look-up tables for constructing the Conceptual Description database for LLM-based composition are also provided, accompanied by visual examples for clearer understanding.
\item \textbf{Automatic Laban Code Detection Workflow} (Sec.~\ref{sec:sup_workflow}): Describes the formulated detection process and the predefined thresholds used for discretization.
\item \textbf{Laban Benchmark} (Sec.~\ref{sec:sup_benchmark}): Provides the mathematical definition of the Laban metric, details on the construction of the HumanML3D-Laban dataset, and illustrative examples.
\item \textbf{Experiment Details} (Sec.~\ref{sec:sup_exp}): Includes training setups, implementation specifics of the proposed models, and prompts used for LLM-based composition.
\item \textbf{Experimental Results} (Sec.~\ref{sec:sup_results}): Extends beyond the main paper by presenting ablation studies, user studies, and qualitative comparisons. Supplementary LLM-composed examples demonstrate that LLMs can interpret ambiguous text and compose Laban symbols.
\item \textbf{Limitations and Future Directions} (Sec.~\ref{sec:sup_plan}): Discusses the current limitations of the proposed framework and outlines possible future research directions.
\end{itemize}
The complete code and related data will be released following approval of the publication status.

\begin{figure}[t]
\centering
\includegraphics[width=0.98\linewidth]{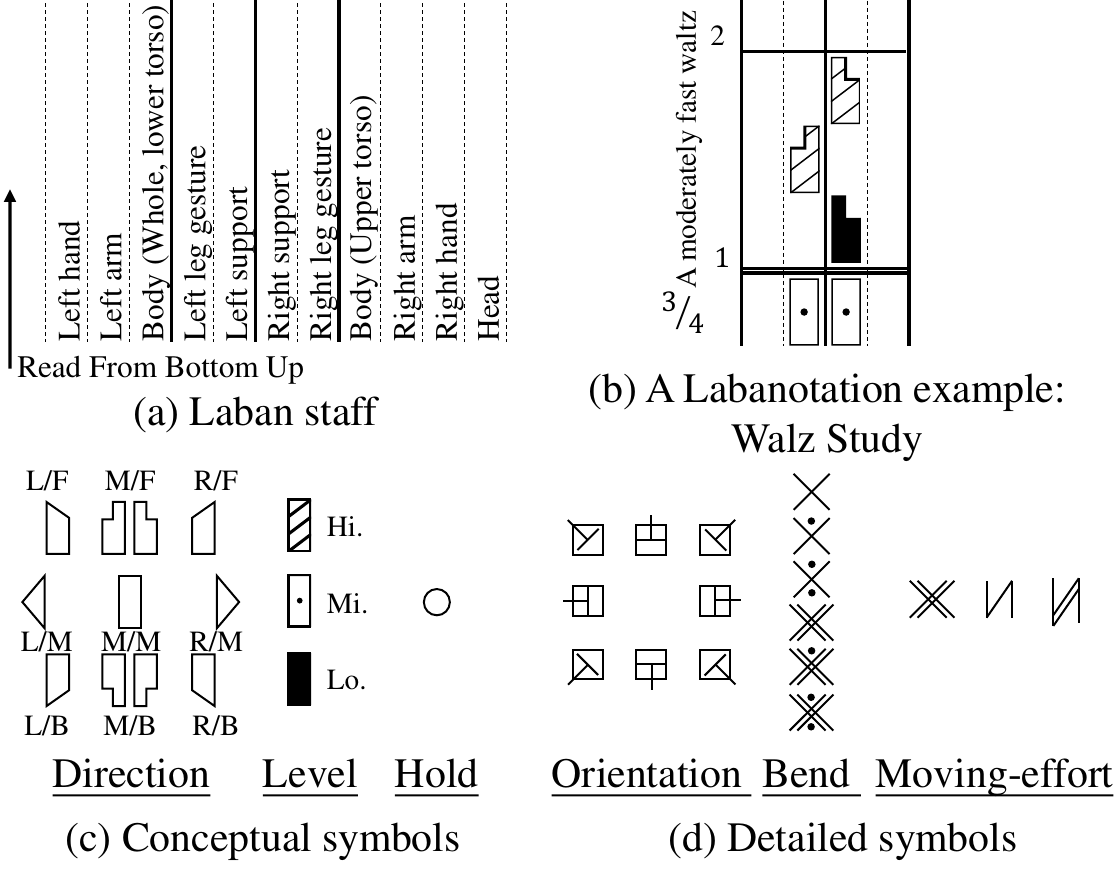}
\caption{\textbf{Preliminary of Labanotation}. (a) Illustration of a {Laban staff}. (b) A segment of a Laban score~\citep{topaz1996elementary}. \textbf{We further extend Labanotation into LabanLite}. LabanLite categorizes Laban symbols into (c) Conceptual symbols, where ``L'', ``R'', ``F'', ``M'', and ``B'' denote left, right, forward, middle, and backward directions. ``Hi.'', ``Mi.'', and ``Lo.'' denote high, middle, and low levels. (d) Detailed symbols, where ``Orientation'', ``Bend'' and ``Moving-effort'' describe the kinematic details. }
\label{fig:lbn_preliminary}
\end{figure}

\begin{figure}[t]
\centering
\includegraphics[width=0.9\linewidth]{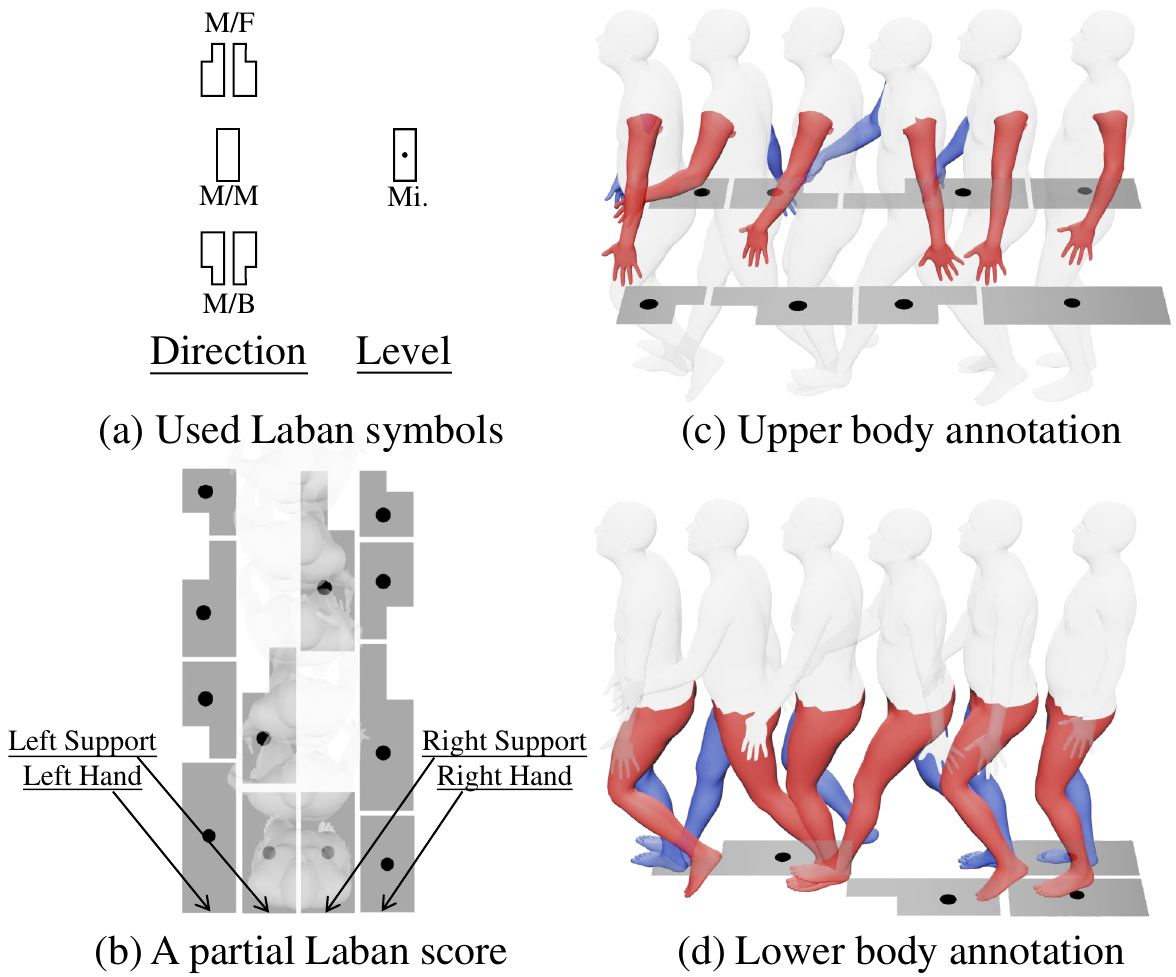}
\caption{Illustration of a partial Laban score. This figure provides a visual explanation of the annotation process for a forward walk movement.}
\label{fig:laban_score_example}
\end{figure}

\section{Details of LabanLite}\label{sec:sup_labanlite}

We begin by introducing Labanotation to provide readers with a quick understanding of Laban theory (Sec.~\ref{sec:lbn_pre}). We then describe LabanLite (Sec.~\ref{sec:lbnlite}), which constitutes our primary contribution.

\subsection{Preliminary of Labanotation}\label{sec:lbn_pre}

Labanotation employs distinctive symbols arranged on a vertical staff to form a Laban score, documenting the spatial positions and temporal sequences of various body parts~\citep{topaz1996elementary}. 
As illustrated in Fig.~\ref{fig:lbn_preliminary}(a), a standard Laban staff consists of 11 columns, each corresponding to a specific body part. Symbols are placed in each column from bottom to top, where the position and sequence reflect the type of movement as well as its start and end times. The length of each symbol indicates the movement duration, rendering Labanotation an event-wise annotation scheme. Choreographers often omit symbols for movements deemed insignificant.
Fig.~\ref{fig:lbn_preliminary}(b) shows a sample score for a waltz study.
Choreographers synchronize the staff with music using bar lines and time signatures. Notably, important movements—such as the right foot's step—are clearly marked with a symbol, while less relevant movements (e.g., the left foot’s status) are often omitted. As a result, some parts of the score contain blank spaces where symbols are absent.

\subsubsection{Quick Tutorial of Labanotation}\label{sec:lbn_tut}

To facilitate understanding, we provide a brief tutorial on Labanotation that serves as a prerequisite for later sections. Here, we explain how to read a Laban score and introduce the annotation protocol, enabling readers to quickly grasp its essential principles.

Labanotation applies specific constraints and simplifications to the symbol sequence for improved human readability, resulting in a set of visualization rules:
\begin{enumerate}
\item Columns that are not of primary interest may be left blank, allowing dancers/performers interpretive freedom for unspecified movements. For example, if a choreographer wishes to focus on a jump, the lower body columns (Left support and Right support) contain relevant symbols, while arm-related columns may be left empty, giving the dancer creative liberty for the upper body.
\item For the Left Support and Right Support columns, a symbol present in one column at a particular time typically means the corresponding position in the other column is left empty.
\item Direction and Level symbols are combined for simplicity.
\item A new Laban symbol appears in a column only when it differs from the preceding symbol; otherwise, it is omitted for clarity and conciseness.
\end{enumerate}

For further clarity, Fig.~\ref{fig:laban_score_example} presents a partial Laban score representing a forward walk. In this example, the Laban staff consists of four columns, from left to right: Left hand, Left support, Right support, and Right hand.
This score abstracts a complex motion sequence into a concise sequence using only four types of symbols: Level ``Mi.''; and Direction ``M/F'', ``M/M'', and ``M/B''. Throughout the walking sequence, the initial state—both feet together and hands naturally lowered—is represented as Direction ``M/M'' with Level ``Mi.'' in all four columns. The left foot then steps forward accompanied by a backward swing of the right arm, which is recorded as a forward-medium symbol set (Direction ``M/F'' and Level ``Mi.'') in the support column, and a backward symbol set (Direction ``M/B'' and Level ``Mi.'') in the hand column. In the final state, the left foot continues forward (Direction M/F and Level ``Mi.''), while the right foot remains behind.

\subsection{LabanLite: A Motion Representation Enhanced over Labanotation}\label{sec:lbnlite}

As discussed in the main paper, to enhance suitability for digital encoding and facilitate composition by large language models (LLMs) in subsequent processes, we introduce LabanLite. LabanLite preserves the expressive power of Labanotation while providing three key enhancements:

\begin{enumerate}
\item \textbf{Symbol classification}: Symbols are classified as either \textit{conceptual} (representing the primary movement structure) or \textit{detailed} (specifying fine-grained attributes);
\item \textbf{Frame-wise annotation}: The event-wise annotation of Labanotation is replaced by frame-wise annotation. Each symbol is converted into a \textit{Laban instance} at every frame, and staff columns are redefined as \textit{body-part groups} with dedicated \textit{attribute fields};
\item \textbf{Structured description}: Each conceptual symbol is accompanied by a strictly formatted textual description (i.e., \texttt{<body-part group> <moving semantic> in <time> seconds})
termed \textit{\textbf{Conceptual Description}}. This standardization enables unambiguous conversion from symbolic event-wise annotations to frame-wise instances via direct textual mapping.
\end{enumerate}

In the following, we first introduce the definitions of conceptual and detailed symbols (Sec.~\ref{sec:lbnlite_symbol}). Next, we describe frame-wise annotation in LabanLite, including Laban instances and their attribute fields within body-part groups (Sec.~\ref{sec:lbnlite_attr}). We then explain the conversion process from event-wise symbol sequences to frame-wise instance sequences (Sec.~\ref{sec:lbnlite_inst}). Finally, we introduce the Conceptual Description and detail the construction of the Conceptual Description database (Sec.~\ref{sec:cd_db}).

\subsubsection{Conceptual and Detailed Laban Symbols}\label{sec:lbnlite_symbol}

In LabanLite, Laban symbols are categorized into two types:
\begin{itemize}
\item \textbf{Conceptual symbols} (\textit{Direction, Level, Hold}): Describe general movement concepts and the structural aspects of motion;
\item \textbf{Detailed symbols} (\textit{Orientation, Bend, Moving-effort}): Capture subtle details of individual body part movements.
\end{itemize}
Tables~\ref{tab:direction},~\ref{tab:level},~\ref{tab:hold},~\ref{tab:orient},~\ref{tab:bend}, and \ref{tab:speed} show the names, graphical appearances, and \textit{partial semantic meanings} of Laban symbols: Direction, Level, Hold, Orientation, Bend, and Moving-effort, respectively. 
We refer to the semantic meanings as ``partial'' because the full meaning of a symbol depends on additional factors such as the body-part group, its duration (length), and its specific position on the Laban staff.
Additionally, for the ``M/F'' and ``M/B'' symbols in the Direction category, each typically has two graphical forms. The specific form is determined by the staff column (i.e., body part): for example, if the ``M/F'' symbol is placed in the ``Left hand'' column, the left-facing form is used; in the ``Right hand'' column, the right-facing form is used.

\subsubsection{Body-Part Group and Movement Attributes}\label{sec:lbnlite_attr}

LabanLite extends the definition of staff columns from Labanotation by introducing body-part groups. Each column now represents a body-part group associated with a set of movement attributes, corresponding to specific sets of Laban symbols.

Importantly, a single body-part group may be represented by multiple staff columns since a group can include multiple body parts, each with distinct attributes and corresponding Laban symbols. This design reflects the constraints of human kinematics—not every body part possesses the same attribute fields. For example, the Level symbol is not applicable to the head column, as the head cannot independently move to arbitrary heights; thus, the Head does not have the Level attribute.

This design enables simultaneous recording of both conceptual and detailed movement information. Detailed definitions of body-part groups and their attributes are provided in Table~\ref{tab:codebook}.

For greater clarity, consider the movement ``wave left hand.'' The relevant body-part group is ``Upper-L,'' which includes the ``Left arm'' and ``Left hand'' staff columns. The Direction, Level, and Hold symbols are assigned to the ``Left hand'' column to indicate the high-level movement concept (e.g., raising up the left arm), while the Bend symbol is placed on the ``Left arm'' column to capture lower-level movement details (e.g., flexing and stretching the left arm). Both symbols are assigned to the body-part group ``Upper-L''.

\begin{figure}[t]
\centering
\includegraphics[width=0.95\linewidth]{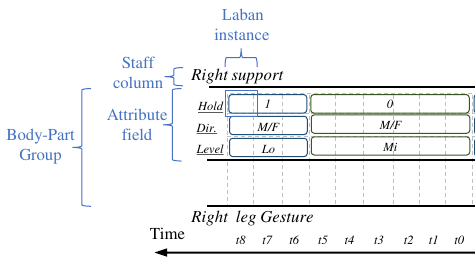}
\caption{Illustration of the relationships among Body-Part Groups, attribute fields, Laban staff columns, and Laban instances.}
\label{fig:relationship}
\end{figure}

\begin{figure}[t]
\centering
\includegraphics[width=1\linewidth]{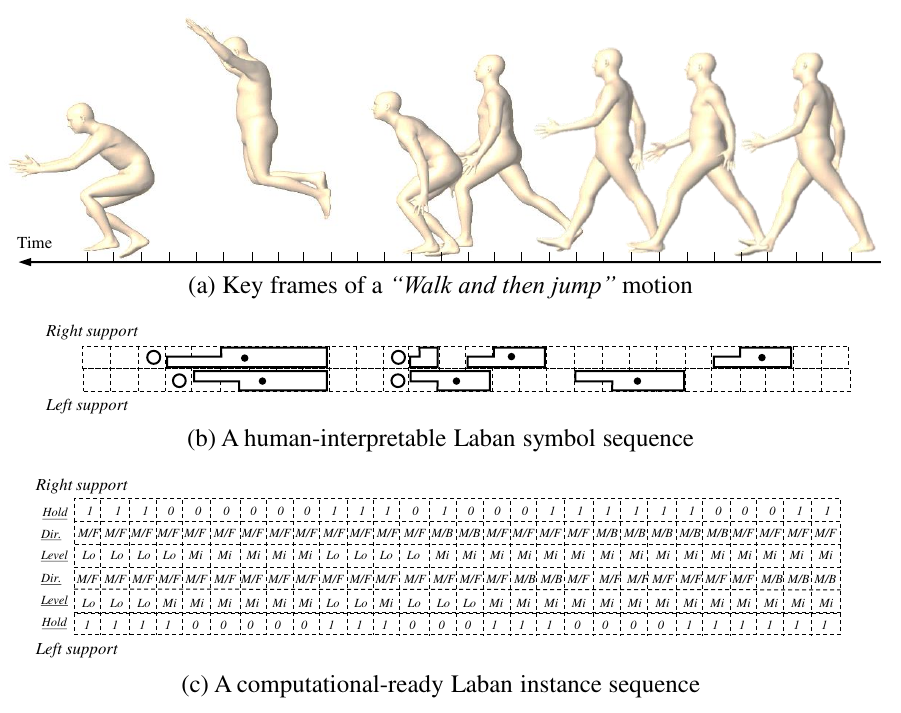}
\caption{Visual comparison between a human-interpretable symbol sequence and a computational-ready instance sequence. The illustrated Laban staff columns correspond to Left support and Right support.}
\label{fig:symbol_instance}
\end{figure}

\subsubsection{Symbol Instantiation: From Event-wise to Frame-wise}\label{sec:lbnlite_inst}
Traditional Labanotation is fundamentally an event-wise annotation and motion analysis system: a placed Laban symbol indicates the corresponding body part's movement, along with its onset and duration. However, modeling temporal coordination across multiple body parts becomes challenging with event-wise symbol sequences. This is because symbols with differing onset times and durations create asynchronous event streams that are difficult to align and analyze together, limiting their suitability for computation-driven tasks such as automatic decoding or generation.

To make the symbol sequence computationally tractable, we define a \textit{Laban instance} as a unique pairing of a symbol and a Body-Part Group at a specific frame. This enables a frame-wise annotation, where explicit values are assigned to every attribute for each body-part group in every frame. The relationships among Laban symbols, staff columns, Laban instances, and Body-Part Groups are illustrated in Fig.~\ref{fig:relationship}. 

Specifically, as described in the main paper, the instantiation process is formulated as follows. 
Given an input motion sequence of $T$ frames, $X = \{x_t\}_{t=1}^{T}$, we propose an Automatic Symbol Detection Workflow $\mathcal{F}$, which converts $X$ into a sequence of instances $S=\mathcal{F}(X)$, where:
\begin{equation}
S=\{s_{t}^{i,j}~|~t\in[1,T],~i\in[1,A_j]\,~j\in[1,G] \}.
\label{eq.instance}
\end{equation}
Here, $G$ denotes the total number of Body‐Part Groups, and $A_j$ represents the attribute field dimension for Group $j$. Note that different Body‐Part Groups may be associated with distinct sets of motion attribute fields, due to variations in their movement ranges and functional roles.

\paragraph{Discussion: Human-Interpretable vs. Computational-Ready.}
As previously discussed, Labanotation applies specific rules and simplifications to improve human readability, often leaving columns or time intervals blank to keep the score clear and concise. In contrast, computational applications such as decoding motion from Laban codes (described in Sec. 3.1.2 of the main paper) or generating new codes require instantiating every symbol at every frame—ensuring each attribute field for every Body-Part Group has \textbf{\textit{an explicit, unambiguous value, without omissions}}.

To address these needs, we introduce the concept of \textit{Laban instances}, which makes the Laban symbol sequence fully dense and computationally ready. 
Fig.~\ref{fig:symbol_instance} compares human-interpretable event-wise sequences and computational-ready frame-wise Laban instance sequences for lower body movement, highlighting the differences between a traditional Laban score and our proposed approach.

\subsubsection{Conceptual Description and Conceptual Description Database}\label{sec:cd_db}

As defined in the main paper (Sec.~3.1.1), a conceptual symbol can be mapped to a Conceptual Description with the format ``\texttt{<body-part group> <moving semantic> in <time> seconds}''. We maintain a Conceptual Description database to enable LLMs to compose motion plans at the conceptual level via retrieval-augmented prompting (Sec.~3.3.1). 

Table~\ref{tab:sup_semantic} and Table~\ref{tab:arm_semantic} provide definitions for Conceptual Descriptions. Taking the support body-part group as an example, ``Support-L'' represents the left support and ``Support-R'' represents the right support; here, we focus on the right support. The moving semantics for this group are defined in Table~\ref{tab:sup_semantic}, such as ``steps to right.'' Through the Automatic Laban Code Detection Workflow, we can obtain the duration of this movement (e.g., 2 seconds). Thus, the corresponding Conceptual Description is ``\texttt{Support Right steps to right in 2 seconds}''.

This description can be unambiguously mapped back to its conceptual symbol, providing LLMs with an explicit interface to interpret and synthesize motion. By modifying the conceptual description, LLMs can alter the corresponding Laban symbols and thereby revise the motion plan. Owing to the standardized format of conceptual descriptions, these mappings are precise and facilitate effective symbolic motion synthesis from textual input. In this supplementary material, Sec.~\ref{sec:lbn_cd} presents examples of LLM-edited conceptual results, illustrating two key capabilities: (1) LLMs can edit existing conceptual symbols while preserving the temporal structure, and (2) even when provided with ambiguous descriptions, LLMs are able to compose symbols that closely align with the user's intended meaning.

\begin{figure}[t]
\centering
\includegraphics[width=0.98\linewidth]{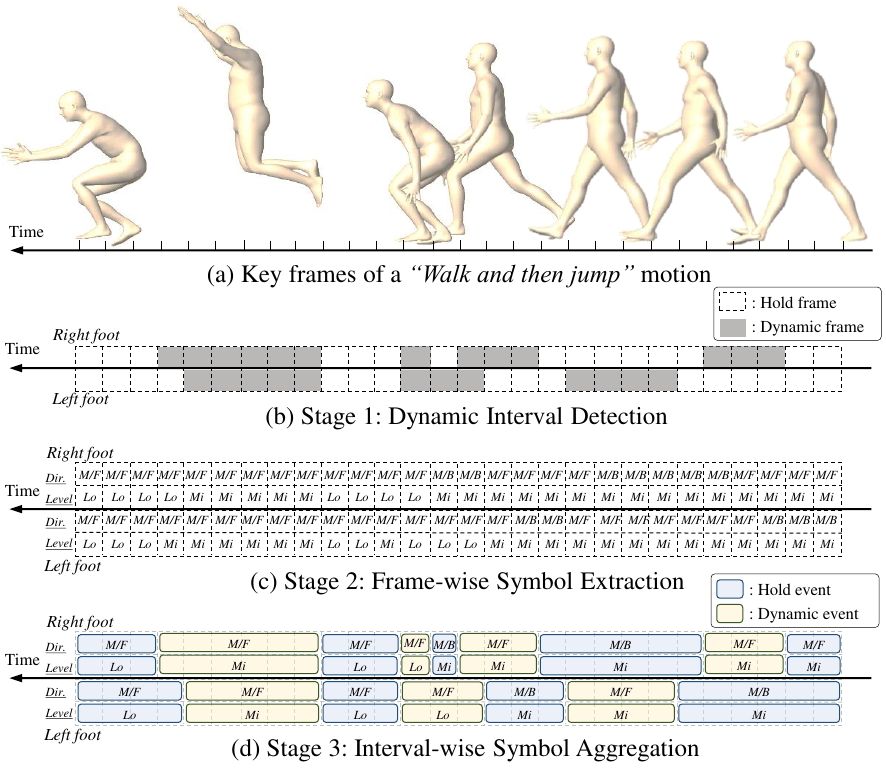}
\caption{Illustration of extracting lower-body {conceptual symbols}: (a) Given a motion sequence; (b) we segment the sequence into dynamic and hold intervals by computing foot velocity; (c) in parallel, frame-wise Laban symbols are identified; (d) for each interval, aggregated frame-wise symbols yield the most representative symbol for the interval.}
\label{fig:lower_lbn}
\end{figure}

\section{Automatic Laban Code Detection Workflow}\label{sec:sup_workflow}

As described in the main paper, the proposed workflow consists of three steps: (1) Dynamic Interval Segmentation, which identifies dynamic and hold intervals; (2) Frame-wise Symbol Extraction, which converts the pose of each frame into corresponding Laban symbols; and (3) Interval-wise Symbol Aggregation, which selects the most representative Laban symbol for each interval. 
As illustrated in Fig.~\ref{fig:lower_lbn}, we demonstrate an example of the lower body symbol detection process. 
Here, we provide a detailed formulation of the Frame-wise Symbol Extraction step, including the associated predefined thresholds.

Specifically, all motion sequences are parameterised using the SMPL model~\citep{SMPL-X:2019} without facial and finger key joints. Key joints such as the left/right hand, elbow, shoulder, hip, knee, foot, pelvis, and spine2 are extracted to compute the corresponding Laban symbol sequences.

\subsection{Details of Frame-wise Symbol Extraction}

\paragraph{Direction and Level symbols.}

To ensure consistency, all body motions are transformed into a canonical space. This is achieved by fixing the root (pelvis) of each pose at the origin, aligning the triangular plane formed by the pelvis, right leg, and left leg key joints with the xz-plane, and orienting the body to face the negative y-axis.
For each key joint (left/right hand, elbow, foot, knee), the $L_2$-norm distance to the pelvis is projected onto the x, y, and z planes, denoted as $a$, $b$, and $c$, respectively. These projected distances are compared against predefined thresholds to assign Direction and Level symbols for different body part groups (Support-L, Support-R, Upper-R, Upper-L).

\begin{figure}[t]
\centering
\includegraphics[width=0.5\linewidth]{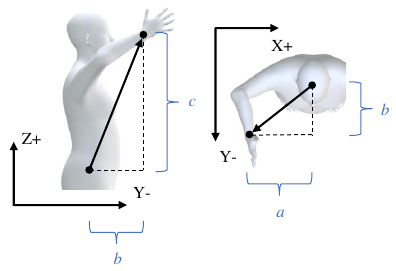}
\caption{Illustration of extracting Direction and Level symbols for the right hand, in a canonical space.}
\label{fig:lbn_howto}
\end{figure}

For the lower body, the thresholds are as follows:
\begin{itemize}
\item Direction (x-axis): $a < -0.1$ is assigned ``R''; $a > 0.3$ is ``L''; otherwise, ``M''.
\item Direction (y-axis): $b < -0.15$ is ``F''; $b > -0.05$ is ``B''; otherwise, ``M''.
\item Level (z-axis): $0 > c > -0.8$ is ``Lo.''; $c > 0$ is ``Hi.''; otherwise, ``Mi.''.
\end{itemize}
For the upper body:
\begin{itemize}
\item Direction (x-axis): $a < -0.1$ is ``R''; $a > 0.3$ is ``L''; otherwise, ``M''.
\item Direction (y-axis): $b < -0.2$ is ``F''; $b > 0.1$ is ``B''; otherwise, ``M''.
\item Level (z-axis): $c < -0.2$ is ``Lo.''; $c > 0.1$ is ``Hi.''; otherwise, ``Mi.''.
\end{itemize}
Figure~\ref{fig:lbn_howto} illustrates the distance calculation for the left hand as an example.

\paragraph{Hold symbol.}
The velocity magnitude of each hand and foot is analysed. Local maxima in the x, y, and z velocity components are identified to determine the turning points of the wrist’s three-dimensional trajectory. Frames in which the velocity falls below a predefined threshold are labelled as ``hold''. The threshold is set to 0.015 for the feet and 0.0005 for the hands.

\paragraph{Bend symbol.}
Euler angles between adjacent body segments are computed and discretised into six intervals, each spanning $30^\circ$.

\paragraph{Orientation symbol.}
The facing orientation is determined by calculating the angle between the line connecting the hip key joints and the negative y-axis. The resulting angle is quantised into eight discrete directions, each spanning $45^\circ$.

\paragraph{Moving-effort symbol.}
The global absolute velocity of the pelvis is computed and discretised using predefined intervals. The velocity components on the $xy$- and $yz$-planes are assigned to one of five speed categories: 0 (very slow), 1 (slow), 2 (normal), 3 (fast), and 4 (very fast), according to the following rules: For both horizontal and vertical velocity components:
\begin{itemize}
\item $0.1 < v \leq 0.5$: label 1;
\item $0.5 < v \leq 1.0$: label 2;
\item $1.0 < v \leq 2.0$: label 3;
\item $v > 2.0$: label 4;
\item $v \leq 0.1$: label 0.
\end{itemize}

\section{Details of Laban Benchmark}\label{sec:sup_benchmark}

\subsection{Laban Metrics}\label{sec:sup_laban_metrics}

Given the ground-truth Laban instance sequence $S$ and the generated Laban instance sequence $\hat{S}$, the proposed three metrics, including Semantic Alignment (SMT), Temporal Alignment (TMP) and Harmonious Alignment (HMN), calculate the similarity between $S$ and $\hat{S}$ using the Longest Common Subsequence (LCS) length. 

Specifically, for SMT and TMP, we measure the Left/Right hand and foot body parts, while for HMN, we measure the body part pairs of {[Left hand, Right hand]}, {[Left foot, Right foot]}, {[Left hand, Left foot]}, {[Left hand, Right foot]}, {[Right hand, Left foot]}, and {[Right hand, Right foot]}. In the main paper, we report the average HMN scores of {[Left hand, Left foot]}, {[Left hand, Right foot]}, {[Right hand, Left foot]}, and {[Right hand, Right foot]} pairs due to the table scale limit. 

Based on Eq.~\ref{eq.instance}, a Laban instance sequence is defined by $S=\{s_t^{i,j}\}$ where $t$, $i$, $j$ denote the frame index, attribute index, and Body-Part Group index, respectively. 
We combine the same attribute from each Body-Part Group across frames to form a duration-ignored Laban instance sequence $\tilde{S}$: 
\begin{equation}
\tilde{S} = \{ \tilde{s}_n^{i,j}~|~n\in[1,N_{i,j}],~i\in[1,A_j]\,~j\in[1,G] \},
\end{equation}
where $n$ denote the duration-ignored Laban instance index and $N_{i,j}$ represents the total instance number of $i$-th attribute and $j$-th Body-Part Group. 

To calculate the selected body part's Laban metric, we fix the attribute index and Body-Part Group index, considering the subset $S_{i^{\star}, j^{\star}} = \{ s_t^{i^{\star}, j^{\star}} \mid t \in [1, T] \}$ and the duration-ignored subset $\tilde{S}_{i^{\star}, j^{\star}} = \{ \tilde{s}_n^{i^{\star}, j^{\star}} \mid n \in [1, N_{i,j}] \}$.

\paragraph{Semantic Alignment (SMT)} evaluates the similarity of inter-body part Laban instances while disregarding their durations. 
Given an attribute index-fixed, Body-Part Group-fixed duration-ignored subset $\tilde{S}_{i^{\star}, j^{\star}}$ and its generation $\hat{\tilde{S}}_{i^{\star}, j^{\star}}$, we formulate the LCS, computing under dynamic programming as follows:
\begin{equation}
f(u,v)=\left\{\begin{matrix}
0 \;\;\; \textup{ if }u=0\textup{ or }v=0, \\
f(u-1,v-1)+1 \;\;\; \textup{ if } \tilde{s}_u^{i^{\star},j^{\star}}=\hat{\tilde{s}}_v^{i^{\star},j^{\star}}, \\
\mathrm{max}(f(u-1,v), f(u, v-1)) \;\textup{otherwise},
\end{matrix}\right.
\label{eq.lcs_1}
\end{equation}
Where $u$ and $v$ denote the index of the duration-ignored Laban instance sequence, i.e., $u, v \in N_{i^{\star},j^{\star}}$.
Such that, the SMT between $\tilde{S}_{i^{\star},j^{\star}}$ and $\hat{\tilde{S}}_{i^{\star},j^{\star}}$ is calculated by the normalised length of the LCS:
\begin{equation}
\mathrm{Sim}_{\mathrm{SMT}}(\tilde{S}_{i^{\star},j^{\star}}, \hat{\tilde{S}}_{i^{\star},j^{\star}}) = \frac{f(N_{i^{\star},j^{\star}},\hat{N}_{i^{\star},j^{\star}})}{\max(N_{i^{\star},j^{\star}}, \hat{N}_{i^{\star},j^{\star}})}.
\end{equation}

\paragraph{Temporal Alignment (TMP)} evaluates the similarity of each inter-body-part Laban instance while considering each symbol's duration, to ensure that not only the types but also the temporal extents of the motions are consistent between the ground truth and the generated sequence. To account for the duration of each symbol, the inputs are changed to Laban instance sequences ${S}_{i^{\star},j^{\star}}$ and $\hat{{S}}_{i^{\star},j^{\star}}$, and we modify Eq.~\ref{eq.lcs_1} as:
\begin{equation}
g(u,v)=\left\{\begin{matrix}
0 \;\;\; \textup{ if }u=0\textup{ or }v=0, \\
g(u-1,v-1)+1 \;\;\; \textup{ if } {s}_u^{i^{\star},j^{\star}}=\hat{{s}}_v^{i^{\star},j^{\star}}, \\
\mathrm{max}(g(u-1,v), g(u, v-1)) \;\textup{otherwise},
\end{matrix}\right.
\end{equation}
where $u$ and $v$ denote the frame index, i.e., $u,v\in T$.
The Temporal Alignment score is then:
\begin{equation}
\mathrm{Sim}_{\mathrm{TMP}}({S}_{i^{\star},j^{\star}}, \hat{{S}}_{i^{\star},j^{\star}}) = \frac{g(T,\hat{T})}{\mathrm{max}(T,\hat{T})}.
\end{equation}

\paragraph{Harmonious Alignment (HMN)} evaluates the synchronous occurrence of Laban symbols across pairs of body parts. Given a body part pair specified by $[(i_1,j_1), (i_2,j_2)]$, and their corresponding duration-ignored Laban instance sequences $\tilde{S}_{i_1,j_1}$, $\tilde{S}_{i_2, j_2}$ for the ground truth, and $\hat{\tilde{S}}_{i_1,j_1}$ and $\hat{\tilde{S}}_{i_2,j_2}$ for the generated motion, we proceed as follows. For each instance $ \tilde{s}_u^{i_1,j_1} \in \tilde{S}_{i_1,j_1}$, where $1 \leq u \leq N_{i_1,j_1} $, we identify the instance $ \tilde{s}_v^{i_2,j_2} \in \tilde{S}_{i_2,j_2}$, where $1 \leq v \leq N_{i_2,j_2} $, whose temporal span overlaps with $\tilde{s}_u^{i_1,j_1}$. If the intersection over union of their durations exceeds 50\%, we consider $\tilde{s}_u^{i_1,j_1}$ and $\tilde{s}_v^{i_2,j_2}$ to occur synchronously. These synchronously occurring symbol pairs are collected into a sequence of combined symbol tuples, denoted by $\mathbb{S}_{j_1, j_2} =\{ (\tilde{s}_u^{i_1,j_1}, \tilde{s}_v^{i_2,j_2}) \}$, and similarly for the generated sequence $\hat{\mathbb{S}}_{j_1, j_2}$. Finally, the HMN similarity is computed as:
\begin{equation}
\mathrm{Sim}_{\mathrm{HMN}}(\mathbb{S}_{j_1, j_2}, \hat{\mathbb{S}}_{j_1, j_2}) = \mathrm{Sim}_{\mathrm{SMT}}(\mathbb{S}_{j_1, j_2}, \hat{\mathbb{S}}_{j_1, j_2}).
\end{equation}

\begin{figure}[t]
\centering
\includegraphics[width=0.85\linewidth]{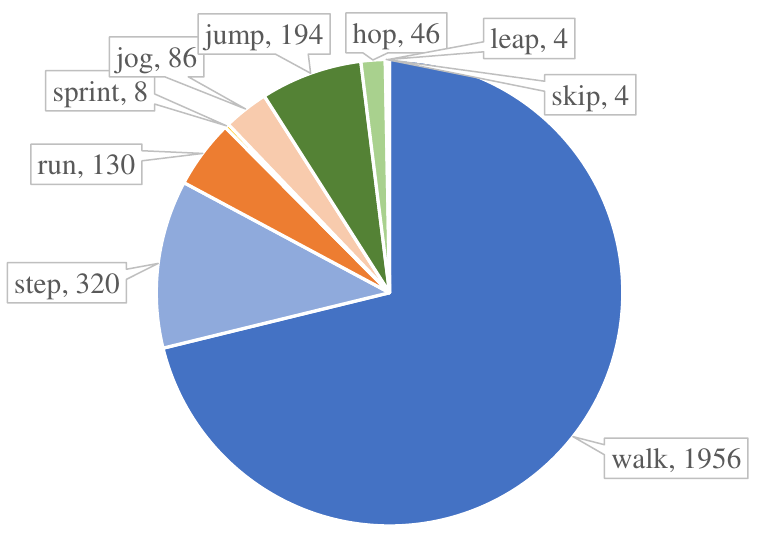}
\caption{Distribution of action classes in the HumanML3D-Laban dataset.}
\label{fig:loco_act}
\end{figure}

\begin{figure}[t]
\centering
\includegraphics[width=0.85\linewidth]{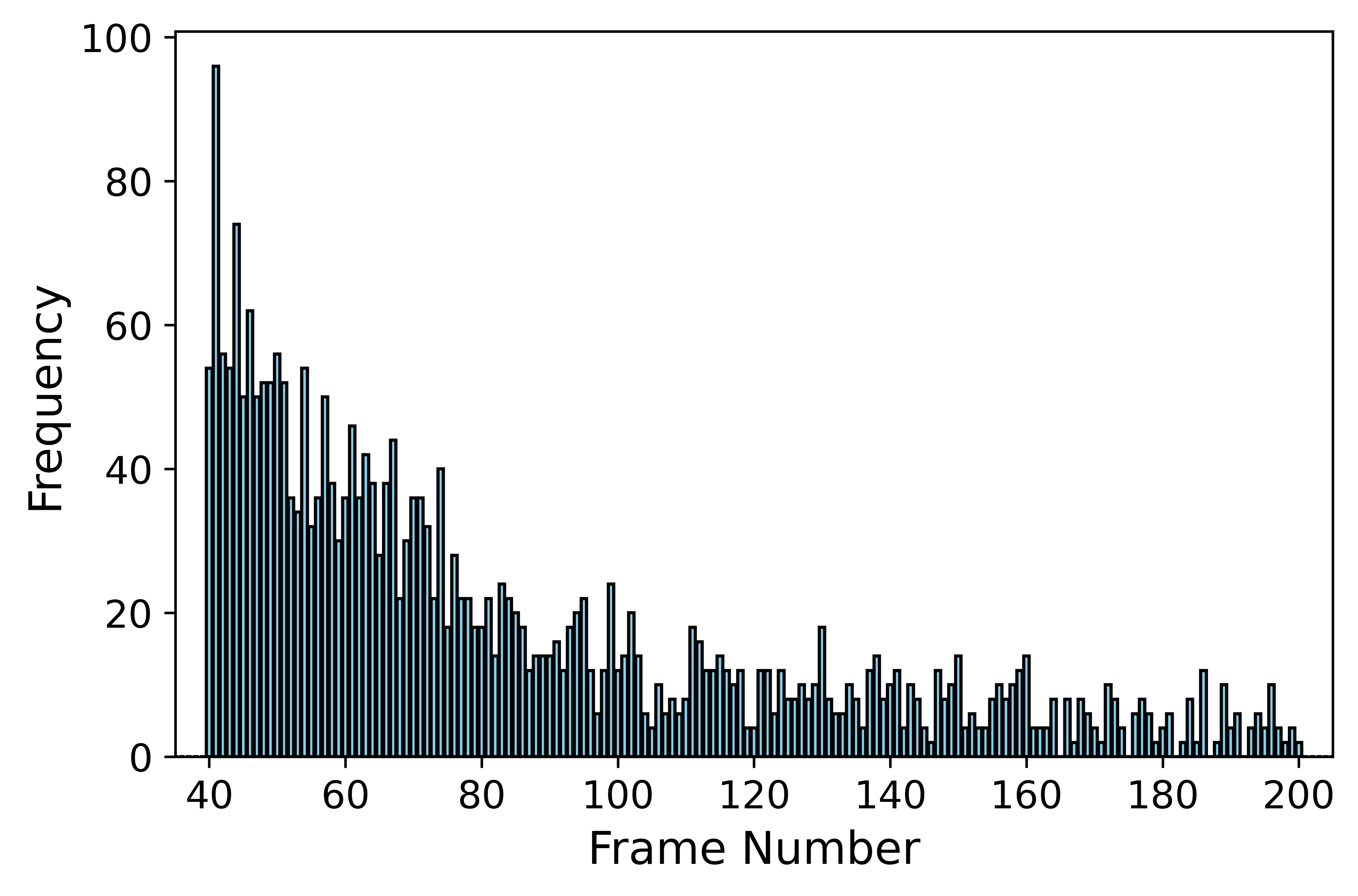}
\caption{Distribution of the number of frames per motion sequence in the HumanML3D-Laban dataset.}
\label{fig:loco_frame}
\end{figure}

\begin{figure}[t]
\centering
\includegraphics[width=0.98\linewidth]{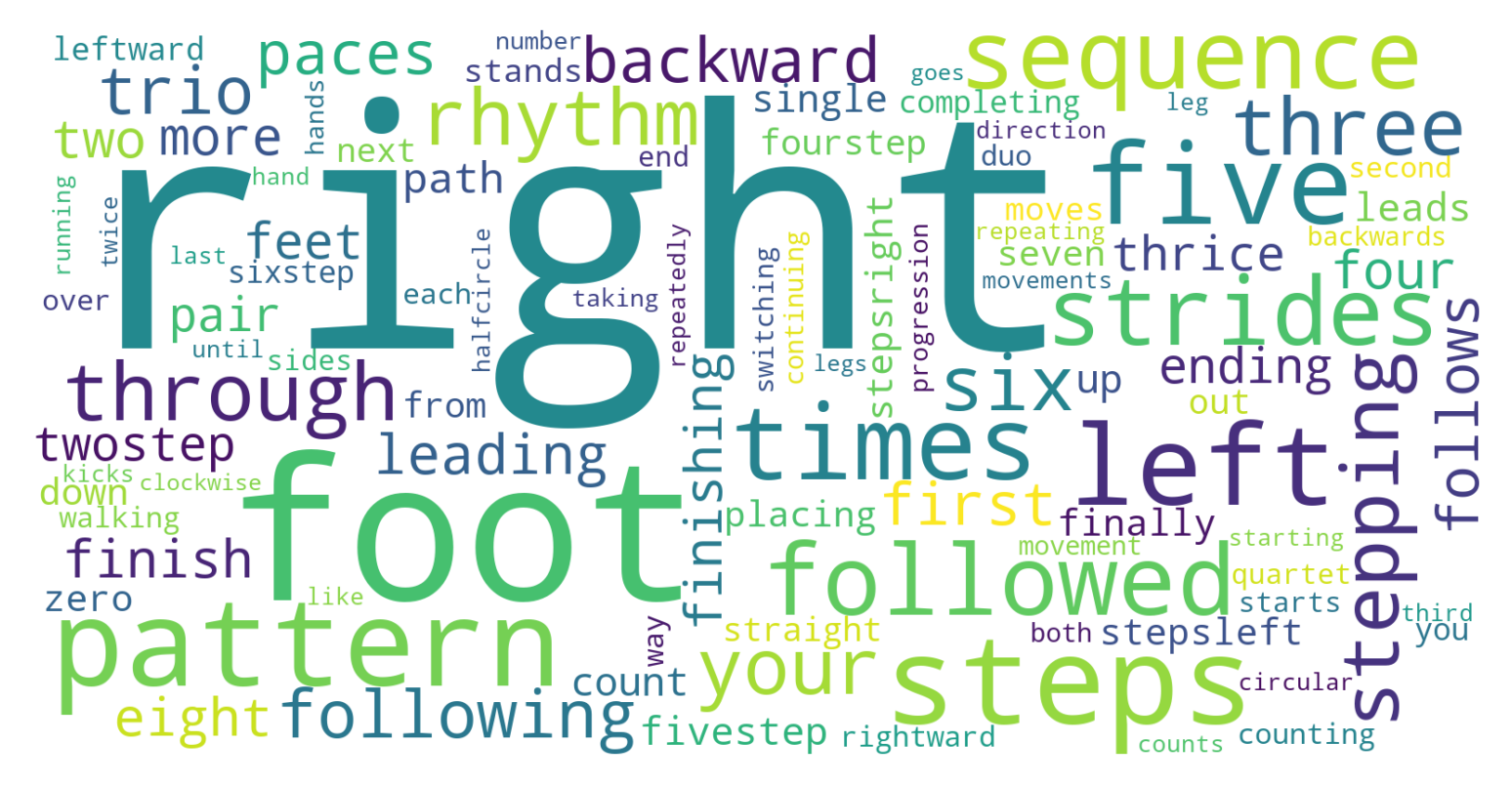}
\caption{\textbf{WordCloud} of the most frequent words in the HumanML3D-Laban descriptions, generated using the Python package \texttt{wordcloud}~\citep{wordcloud}.}
\label{fig:loco_anno}
\end{figure}

\subsection{HumanML3D-Laban: Motion-Laban-Text Paired Dataset}

This section presents the construction and characteristics of our HumanML3D-Laban dataset, including statistics on motion lengths, instructional text descriptions, and examples of text-motion pairs. Motion sequences were selected from the HumanML3D~\citep{guo2022humanml3d} dataset, covering mostly locomotions such as walking, running, stepping, and jumping. Following the annotation approach of BABEL-TEACH~\citep{athanasiou2022teach}, these actions were further decomposed into atomic actions and annotated accordingly.

\paragraph{Semi-automatic Annotation.}
The annotation process was conducted in a semi-automatic manner. Initially, we manually inspected rendered HumanML3D motion sequences to extract detailed motion information, such as the number of steps, their body part sequence, and the action label (e.g., Walk forward: a three-step walk with the order: right, left, right). These details were stored in a JSON format with keys including ``step number,'' ``step order,'' and ``action label.'' Subsequently, we utilised GPT-4.1~\citep{achiam2023gpt} to generate natural language descriptions by integrating the motion details. The prompt used for this step is shown in the second entry of Table~\ref{tab:prompt}. Finally, we paired the rephrased instructional text descriptions with their corresponding motions, forming the motion-laban-text pairs of the HumanML3D-Laban dataset. In accordance with the HumanML3D~\citep{guo2022humanml3d} evaluation protocol, we restricted the motion sequence lengths to between 40 and 200 frames.

\paragraph{Examples of motion details and motion-laban-text pairs.}
In the supplementary materials, we provide the complete set of instructional text descriptions that have been rephrased by an LLM. Specifically, within the ``annotations'' folder, each text file corresponds to a single motion instance and contains its associated instructional text description. For each motion instance, we used GPT-4.1 to generate five distinct rephrasings of the description, resulting in five different annotations per text file.

\paragraph{Statistics.}
Following the procedure in \citet{guo2022humanml3d}, all HumanML3D motion sequences were downsampled from 120 FPS to 20 FPS. We further filtered out motion sequences with fewer than 40 or more than 200 frames. As a result, the final dataset comprises 2,748 text-motion pairs, corresponding to approximately 18 hours of human motion data. Fig~\ref{fig:loco_act} illustrates the distribution of action labels, while Fig.~\ref{fig:loco_frame} shows the distribution of frame lengths. Fig.~\ref{fig:loco_anno} provides an overview of the most frequent words in the instructional text descriptions, drawn from a vocabulary of 1,220 unique words.

\section{Experiment Details}\label{sec:sup_exp}

\begin{table}[t]
\centering
\caption{The detailed structure of the Decoder in the Laban-Motion Encoder-Decoder module.}
\begin{tabular}{lc}
\toprule
Encoder & Num. \\ 
\midrule
In Dim. & 512 \\
Feat. Dim. & 512 \\
Depth & 8 \\
Head & 8 \\
Head Dim. & 64 \\
FFN Dim. & 1024 \\
Out Dim. & 512 \\
\bottomrule
\end{tabular}
\label{table:decoder}
\end{table}

\subsection{Implementation Details}

\paragraph{Network Architecture.}
For the Laban-Motion Encoder-Decoder, the Encoder operates as a rule-based, non-learning process, while the Decoder is implemented as a conventional Transformer-based decoder. The Decoder incorporates standard Attention modules~\citep{vaswani2017attention}, including Multiheaded Self-Attention blocks (MSAs), and Feed-Forward Network blocks (FFNs), with Layernorm (LN) applied before each module. The detailed architecture of the Decoder is summarised in Table~\ref{table:decoder}, where ``Head'' and ``Head Dim.'' refer to the number of attention heads and the dimensionality of the features in the MSA blocks, respectively.   
For the Motion Generator, we adopt the configuration from \citet{zhang2023t2mgpt,huang2024controllable} for the Motion Generator. Specifically, a linear layer first projects the Laban code sequences, after which positional encoding is applied. The resulting sequence is then processed by a decoder-only Transformer comprising causal self-attention blocks.

\paragraph{Training settings.}
The Decoder and Generator are trained with AdamW optimiser (learning rate $1 \times 10^{-4}$, batch size 512). The Decoder is trained for 200k iterations, and the Generator is trained for 100k iterations. 
All experiments are conducted on a workstation equipped with 1 Intel Xeon Gold 6438Y+ CPU and 1 NVIDIA L40 GPU.
We select the checkpoint with the lowest FID on validation for final evaluation.

\paragraph{Testing-time settings.} 
Recall that, during all testing-time inference procedures, the first stage of our generation framework utilizes an LLM to compose motion plans at the conceptual level via retrieval-augmented prompting  (see Sec. 3.3.1 in the main paper). Specifically, the retrieval pool exclusively consists of motion-Laban annotations from the training set; no Laban information from the evaluation set is disclosed to the LLMs. Consequently, at evaluation, the sequences of conceptual symbols are generated entirely by the LLM, relying solely on the motion information and textual descriptions provided in the training set, as well as user input textual descriptions. The LLM then reasons over the alignment between textual concepts and symbolic motion patterns to generate a novel sequence of conceptual symbols based on the input text.

\paragraph{Hyperparameters.} 
We adopt the motion feature extractor~\citep{guo2022humanml3d} to convert HumanML3D-Laban, HumanML3D, and KIT-ML motions into features of dimensions 263, 263, and 261. HumanML3D-Laban and HumanML3D share the same evaluator to get text and motion embeddings. Following \citep{zhang2023t2mgpt}, we set $\lambda=0.5$. Laban codebook contains 37 Laban categories and 158 distinct codes, with a size of $158 \times 512$. We employ two commercial LLM models named GPT-4 mini ($\texttt{gpt-4.1-mini-2025-04-14}$) and GPT-4 ($\texttt{gpt-4.1-2025-04-14}$) from \citet{achiam2023gpt}, two open-source LLM models named Qwen3 (\textit{qwen3-32b}) from \citet{qwen3} and DeepSeekV3 (\textit{deepseek-v3}) from \citet{deepseekv3} for composition. We use CLIP~\citep{radford2021clip} ($\texttt{ViT-B/32}$) for text encoding.

\subsection{Related to Large Language Models}
In the main paper, LLMs are utilised to: (1) autonomously plan and compose Laban code sequences through explicit and interpretable symbolic reasoning (Sec.~\ref{sec:highlvl_plan}); and (2) semi-automatically generate instructional text descriptions for the HumanML3D-Laban dataset during annotation (Sec.~\ref{sec:txt_lbnbenchmark}). Here, we provide a detailed explanation of the procedures for using LLMs in Laban code sequences composition.

\subsubsection{High-level Symbolic Motion Planning}\label{sec:highlvl_plan}

Recall that we utilize LLMs to compose conceptual Laban symbol sequences through retrieval-augmented prompting for high-level symbolic motion planning. Specifically, the retrieved references (from the Conceptual Description database) describe motions using sequential Conceptual Descriptions (CDs) (Sec.~3.1.1 in the main paper). By semantically understanding these CDs and imitating their expression forms, LLMs are able to compose new CDs tailored to the target textual descriptions. According to the characteristics of LabanLite, the composed CDs can be unambiguously converted to Laban instances, which are then mapped to Laban codes (Sec.~3.1.2 in the main paper), and ultimately generate the corresponding motions via our LLM-Guided Text-Laban-Motion Generator (Sec.~3.3 in the main paper).
Below, we detail the prompt design used for LLM-based symbolic motion composition. Examples of LLM outputs are provided in Sec.~\ref{sec:lbn_cd} of this supplementary material.

The prompt consists of the following components (as formulated in Table~\ref{tab:prompt}): (1) the structure definition of the retrieved CD references, (2) the definition of the CD structure, (3) several retrieved CD references, (4) the target textual description from user input.

\paragraph{Structure Definition of Retrieved CD References.} 
We first define the format of the retrieved CD references. Each reference comprises \texttt{[number]}, \texttt{[Caption]}, \texttt{[Support]}, \texttt{[Left hand]}, and \texttt{[Right hand]}. Here, [number] indicates the reference index, \texttt{[Caption]} provides a high-level semantic description of the motion, and \texttt{[Support]}, \texttt{[Left hand]}, and \texttt{[Right hand]} respectively detail the motions of specific body parts, each in CD format.

\paragraph{Definition of CDs.}
We then specify the format for CDs as ``\texttt{<body-part group>} \texttt{<moving semantic>} in \texttt{<time>} seconds'', along with the complete set of possible CDs. This ensures that the LLM selects from a predefined set rather than inventing new CDs, which could hinder the downstream generation process. To optimize LLM token usage, we simplify the representation by introducing two lookup tables (see Table~\ref{tab:sup_semantic} and Table~\ref{tab:arm_semantic}) for lower and upper human bodies, and use integer indices to denote CDs. With this approach, the LLM outputs the integer indices rather than full text, adhering strictly to the simplified CD definitions. Consequently, the output CDs are presented in the form ``\texttt{(right, 2, 0.25)}'', indicating \texttt{<body-part group>}, \texttt{<moving semantic>}, and \texttt{<time>}, respectively.

\paragraph{Retrieved CD References.}
Following the above CD definition, the simplified retrieved CD references are provided to the LLM.

\paragraph{Target Textual Description.}
Finally, the user's target textual description for the desired motion is also added to the prompt. The LLM composes the new motion by leveraging the high-level caption and the motion details from the retrieved CD references, along with the user's textual input.

\subsubsection{Generate Texts for Laban Benchmark}\label{sec:txt_lbnbenchmark}

In our Laban benchmark (Sec.~4 of the main paper), we construct a paired dataset comprising motion, Laban symbols, and textual instructional descriptions. Each data entry consists of: (1) a motion sequence, (2) its corresponding sequence of Laban symbols, and (3) a textual instructional description. The instructional texts are generated via a semi-automated annotation process.

To create these instructional descriptions, we first manually review the rendered SMPL motion videos and annotate critical motion details, including step count, step sequence, and action labels for each motion. These annotated motion details are then provided as input to an LLM (GPT4.1), which generates natural language instructional descriptions for each motion instance. As illustrated in Table~\ref{tab:prompt} (Prompt \#2), the LLM translates the annotated details into instructional text. For reproducibility and robustness evaluation, the LLM generates five instructional descriptions for each motion. All instructional texts produced by the LLM for the HumanML3D-Laban dataset are included in the supplementary material.

\begin{figure}[t]
\centering
\includegraphics[width=0.8\linewidth]{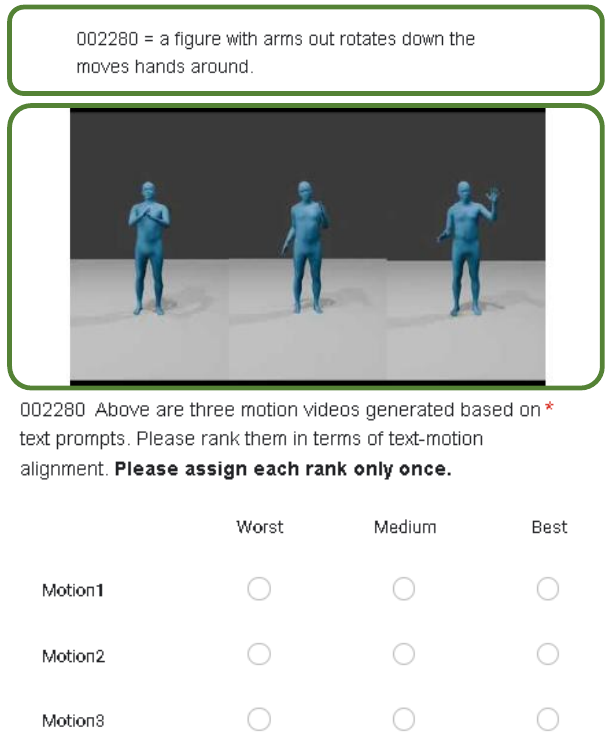}
\caption{Illustration of the user study form.}
\label{fig:user_study_form}
\end{figure}

\begin{figure}[t]
\centering
\includegraphics[width=0.45\linewidth]{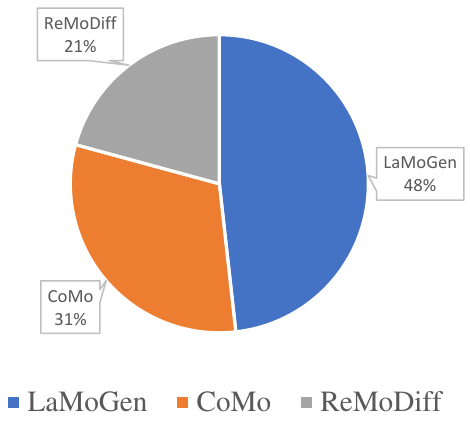}
\caption{Illustration of the user study result.}
\label{fig:user_study_results}
\end{figure}

\section{Experimental Results}\label{sec:sup_results}

\subsection{User Study}

We conducted a user study to evaluate the motion generation quality of LaMoGen (configured with GPT-4.1) in comparison with two state-of-the-art models for fine-grained text-to-motion generation: ReMoDiff~\citep{zhang2023remodiffuse} and CoMo~\citep{huang2024controllable}. Fifteen examples were randomly selected from the HumanML3D test set and rendered into video sequences. For each example, the results from LaMoGen, CoMo, and ReMoDiff were arranged side by side in a randomly determined order, forming a video triplet.

A total of 35 graduate students participated in the evaluation. Each participant was presented with all 15 video triplets and was asked to rank the three generated motions in each triplet as best, second-best, and worst, based on how well they matched the provided instructions, as shown in Fig.~\ref{fig:user_study_form}. Each ranking position was assigned to exactly one method, ensuring mutual exclusivity.

\paragraph{Statistical Analysis.}
For quantitative analysis, we assigned scores of 3, 2, and 1 to the best, second-best, and worst rankings, respectively. The average scores for each method across all motion sequences are presented in Fig.~\ref{fig:user_study_results}. Higher scores indicate stronger user preference.
As shown in Fig.\ref{fig:user_study_results}, LaMoGen achieves the highest average user preference, followed by CoMo and ReMoDiff. Nearly half of the participants favoured motions generated by LaMoGen, indicating that our method provides superior text-motion alignment.

To determine whether the observed differences in user preferences among the three methods are statistically significant, we performed a Friedman test. The results indicate a significant difference in user preference ($\chi^2= 387.380$, $p < 0.05$). Further analysis of the mean ranks demonstrates that LaMoGen achieves the highest mean rank (mean rank = 2.80), followed by CoMo (mean rank = 1.733) and ReMoDiff (mean rank = 1.457). These results suggest that participants consistently preferred LaMoGen over the other methods, with CoMo ranked second and ReMoDiff ranked last.

\subsection{Quantitative Comparisons}

\begin{table}[t]
\setlength{\tabcolsep}{1mm}
\centering\small
\caption{Quantitative comparison for LLM usage cost. The approximate token count per motion sequence for generation and modification is reported. }
\scalebox{0.9}{
\begin{tabular}{rllr}
\toprule
Method & LLM-related Procedure & Objective & $\sim$Token Count (k) \\ 
\midrule
CoMo~\citep{huang2024controllable} & Decompose Texts & Generate & 6.5 \\
& Identify frames & Edit & 51  \\
& Identify body parts and edit & Edit & 53  \\
\cmidrule(lr){2-4} 
& \multicolumn{2}{c}{Total} & 110.5 \\
\midrule
LaMoGen & Compose CDs & Generate & 5 \\
& Modify CDs & Edit & 5 \\ 
\cmidrule(lr){2-4} 
& \multicolumn{2}{c}{Total} & 10  \\
\bottomrule
\end{tabular}}
\label{tab:token_cost}
\end{table}

\subsubsection{LLM Usage Cost}

We report the approximate token count utilized in LLM-based symbolic motion planning per motion sequence. We also compare our approach against CoMo~\citep{huang2024controllable}, a related work that utilizes LLMs for motion generation, in terms of average token count. We reimplement the prompts described in the original paper of CoMo. We estimate the count using OpenAI's token Calculator with GPT4.1. As shown in Table~\ref{tab:token_cost}, our approach requires fewer tokens, indicating lower LLM demand and computational cost. This demonstrates more efficient LLM utilization.

\paragraph{Discussion.}

CoMo quantizes motion at only the pose level, resulting in a 392-code codebook. Each pose is encoded as a combination of body-part codes via the PoseScript~\citep{delmas2024posescript} technique. In CoMo, there are two situations that require LLMs: (1) CoMo generates detailed textual descriptions of target motions; (2) CoMo enables users to edit motions using LLMs. During motion editing, CoMo requires LLM selection and modification from 392 pose codes, which consumes a large number of tokens, making LLM-based editing less efficient and cost-effective.

In contrast, our framework quantizes motions in both pose and temporal domains. Specifically, LabanLite defines 158 Laban codes based on Labanotation and further quantizes each sub-action in the temporal dimension, resulting in a more compact representation. Moreover, our proposed LaMoGen framework utilizes a two-level design that unifies high-level LLM-driven symbolic planning with low-level motion synthesis: LLMs are employed solely at the high-level composition stage, while a trained generator is used for low-level motion completion. This design substantially reduces LLM token consumption and operational cost.

Furthermore, during editing, unlike CoMo, which requires two separate LLM requests—one to identify frames that need modification and another to select and manipulate the corresponding body parts—LaMoGen enables direct modification in a single step. This is made possible by LabanLite's dual quantization in both the pose and temporal domains, allowing LLMs to directly edit sub-actions for each main body part rather than performing frame-wise and part-wise modifications as in CoMo.
Therefore, our framework enables scalable, efficient motion planning while maintaining expressiveness and editability. 

To summarize, this unique framework not only improves LLM token efficiency but also facilitates faster planning, reduces monetary cost, and makes large-scale deployment more feasible.

{
\setlength{\tabcolsep}{1mm}
\begin{table*}[t]
\centering\footnotesize
\caption{Quantitative comparisons on the HumanML3D test set, using the proposed Labanotation-based metrics: Semantic Alignment (SMT), Temporal Alignment (TMP), and Harmonious Alignment (HMN), along with Text-to-Motion metrics: R-precision Top-3 (R@3) and FID. \textbf{Bold} and \underline{underlined} values indicate the best and the second-best performance, respectively.}
\begin{tabular}{rrccccccccccccc}
\toprule
\multicolumn{2}{c}{\multirow{2}{*}{Method}} & \multicolumn{4}{c}{SMT $\uparrow$} & \multicolumn{4}{c}{TMP $\uparrow$} & \multicolumn{3}{c}{HMN $\uparrow$} & \multirow{2}{*}{R@3 $\uparrow$} & \multirow{2}{*}{FID $\downarrow$} \\
\cmidrule(lr){3-6} \cmidrule(lr){7-10} \cmidrule(lr){11-13}
& & supL & supR & armL & armR & supL & supR & armL & armR & arm-arm & arm-sup & sup-sup \\ 
\midrule
\multicolumn{2}{r}{Real data} & - & - & - & - & - & - & - & - & - & - & - & 0.797 & 0.002 \\
\multicolumn{2}{r}{LaMoGen (Dec)} & 0.843 & 0.843 & 0.745 & 0.731 & 0.848 & 0.848 & 0.779 & 0.765 & 0.537 & 0.631 & 0.842 & 0.793 & 0.095 \\
\cmidrule(lr){1-15}
\multicolumn{2}{r}{MDM~\citep{tevet2023mdm}} & 0.413 & 0.413 & 0.276 & 0.315 & 0.372 & 0.372 & 0.369 & 0.308 & 0.177 & 0.284 & 0.502 & 0.611 & 0.544 \\
\multicolumn{2}{r}{ReMoDiff~\citep{zhang2023remodiffuse}} & 0.455 & 0.455 & 0.348 & 0.405 & 0.335 & 0.405 & 0.367 & 0.320 & 0.179 & 0.275 & 0.501 & 0.795 & \textbf{0.103} \\
\multicolumn{2}{r}{MoDiff~\citep{zhang2024motiondiffuse}} & 0.482 & 0.482 & 0.386 & 0.360 & 0.401 & 0.401 & 0.388 & 0.334 & 0.190 & 0.299 & 0.513 & 0.782 & 0.630 \\
\multicolumn{2}{r}{CoMo~\citep{huang2024controllable}} & 0.463 & 0.463 & 0.369 & 0.401 & 0.372 & 0.372 & 0.393 & 0.346 & 0.222 & 0.321 & 0.535 & 0.790 & 0.262 \\
\cmidrule(lr){1-15}
\multirow{4}{*}{\rotatebox{90}{\shortstack{LaMoGen\\Composer}}} & None & 0.467 & 0.467 & 0.353 & 0.381 & 0.374 & 0.374 & 0.338 & 0.363 & 0.218 & 0.313 & 0.529 & 0.755 & 1.091 \\
& GPT4.1mini & 0.437 & 0.437 & 0.331 & 0.354 & 0.517 & 0.517 & 0.463 & 0.445 & 0.270 & 0.371 & 0.561 & 0.779 & 0.561 \\
& GPT4.1 & \underline{0.563} & \underline{0.563} & \underline{0.433} & \underline{0.456} & \underline{0.615} & \underline{0.615} & \underline{0.527} & \underline{0.544} & \underline{0.302} & \underline{0.411} & \underline{0.588} & \underline{0.796} & 0.252 \\
& Human & \textbf{0.690} & \textbf{0.690} & \textbf{0.554} & \textbf{0.534} & \textbf{0.712} & \textbf{0.712} & \textbf{0.623} & \textbf{0.608} & \textbf{0.332} & \textbf{0.451} & \textbf{0.617} & \textbf{0.813} & \underline{0.206} \\
\bottomrule
\end{tabular}
\label{tab:sup_hml3d}
\end{table*}
}

\subsubsection{Inference Time}

We follow the previous work~\citep{huang2024controllable} to assess the inference time using the Average Inference Time per Sentence (AITS). AITS measures the time (in seconds) to generate one motion sequence (not including model loading, dataset preparation, or LLM composition time). We compare our framework against CoMo, where CoMo has a similar architecture, on the HumanML3D test set. As shown in Table~\ref{tab:ab_code_dim}, our approach achieves a lower time cost, demonstrating the compactness of our Laban codebook and the effectiveness of our LaMoGen framework.

\subsubsection{Generation Performance on HumanML3D}

Table~\ref{tab:sup_hml3d} reports the performance results using Laban metrics as well as conventional metrics (R-precision at Top-3, R@3, and FID) on the HumanML3D test set. Consistent with the main paper, we evaluate our model under four high-level composer conditions: (1) \textit{None}, where motion generation is conditioned only on the input text, serving as a baseline for text comprehension; (2) two LLM composers (\textit{GPT4.1mini} and \textit{GPT4.1}), where generation is conditioned on both the text and conceptual cues provided by the LLMs; and (3) \textit{Human}, in which conceptual cues are derived from ground-truth annotations to simulate human composition. We also report the decoded results from ground truth, denoted as \textit{LaMoGen (Dec)}, where both conceptual and detailed Laban codes are provided to the decoder to indicate the upper bound of decoding performance. \textit{Please note: the composer composes conceptual symbols, and the generator generates details symbols. The decoder decodes the combination of conceptual and detail codes.}

Although the HumanML3D text descriptions are not strictly instructional—in other words, they are typically descriptive and do not precisely specify target motions in terms of step count or timing, the performance in the \textit{None} condition is therefore lower than the baselines.

Nevertheless, the use of retrieval-augmented prompting effectively addresses this limitation. For instance, when a user provides a descriptive input such as ``a person walks in a circle'', even though this prompt lacks detailed instructions like the number of steps or timing, the Conceptual Description database is able to retrieve related motions such as ``walk forward'' and ``turn''. The LLM can then compose these motions to generate a ``walk in a circle'' sequence. Furthermore, if an exact example like ``walks in a circle'' exists in the Conceptual Description database, the corresponding conceptual description can be retrieved directly, enabling precise motion synthesis.

As a result, LaMoGen achieves comparable or superior performance on Laban metrics relative to other methods.

\subsection{Qualitative Results}

\subsubsection{Fine-Grained Temporal Structure Modification}\label{sec:lbn_cd}

We further evaluate each method’s ability to model fine-grained temporal structure by modifying the input text with both moving context and vague/precise temporal control. Specifically, we compare our approach with CoMo~\citep{huang2024controllable} and MotionGPT~\citep{jiang2024motiongpt} using the following settings: (1) ``Walk forward'' as the baseline, (2) ``Walk forward slowly'' as the vague speed change, and (3) ``Walk forward in 5 seconds'' as the precise time control.
For each case, we also report the conceptual description composed by the LLM that corresponds to our results (we omit the retrieved conceptual reference, other prompts, and left/right-hand conceptual descriptions due to page limitations).

\begin{figure*}[t]
\centering
\includegraphics[width=0.85\linewidth]{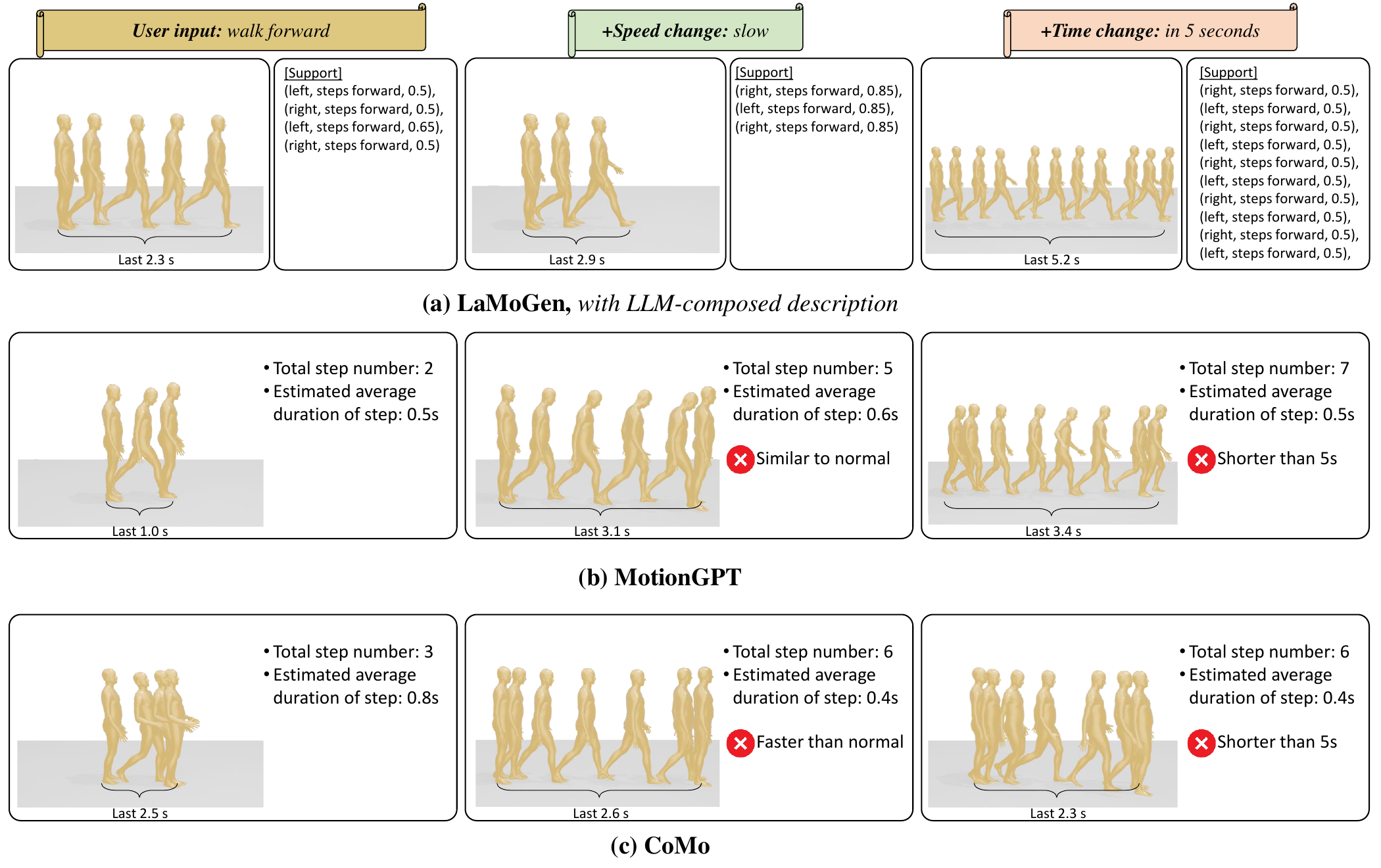}
\caption{Qualitative comparison on fine-grained temporal structure modification. Left: the baseline motion, condition on ``walk forward''. Middle: the motion conditioned on ``walk forward slow''. Right: the motion conditioned on ``walk forward in 5 seconds''. We highlight the step number and the corresponding duration. We calculate the average step duration to assess the degree of the speed change.}
\label{fig:sup_mot_time}
\end{figure*}

Fig.~\ref{fig:sup_mot_time} presents the generated motions of our framework alongside those from other compared methods.  For our approach, we report the corresponding LLM-composed textual descriptions. We measured the total duration of all generated motions and manually selected the keyframes at foot contact to visually indicate the number of steps. This allows us to estimate the average duration of each step and thus assess the degree of motion speed change. It can be observed that CoMo and MotionGPT struggle to accurately modify the temporal structure: CoMo does not effectively adjust the motion speed, and both CoMo and MotionGPT are unable to strictly execute motions according to the user-specified duration.

In contrast, our framework leverages the compositional capabilities of LLMs to explicitly determine the type and order of sub-actions, as well as to set the duration for each sub-action. For example, in the ``Walk forward in 5 seconds'' scenario, our method generates ten steps, allocating 0.5 seconds to each step to support long-term motion. The LLM will adaptively increase the number of steps as needed in order to fulfill the specified duration requirement. These results demonstrate that, thanks to the integration of LabanLite and the compositional reasoning capabilities of LLMs, our framework offers superior modeling of fine-grained temporal structure compared to existing approaches.

\begin{figure*}[t]
\centering
\includegraphics[width=0.85\linewidth]{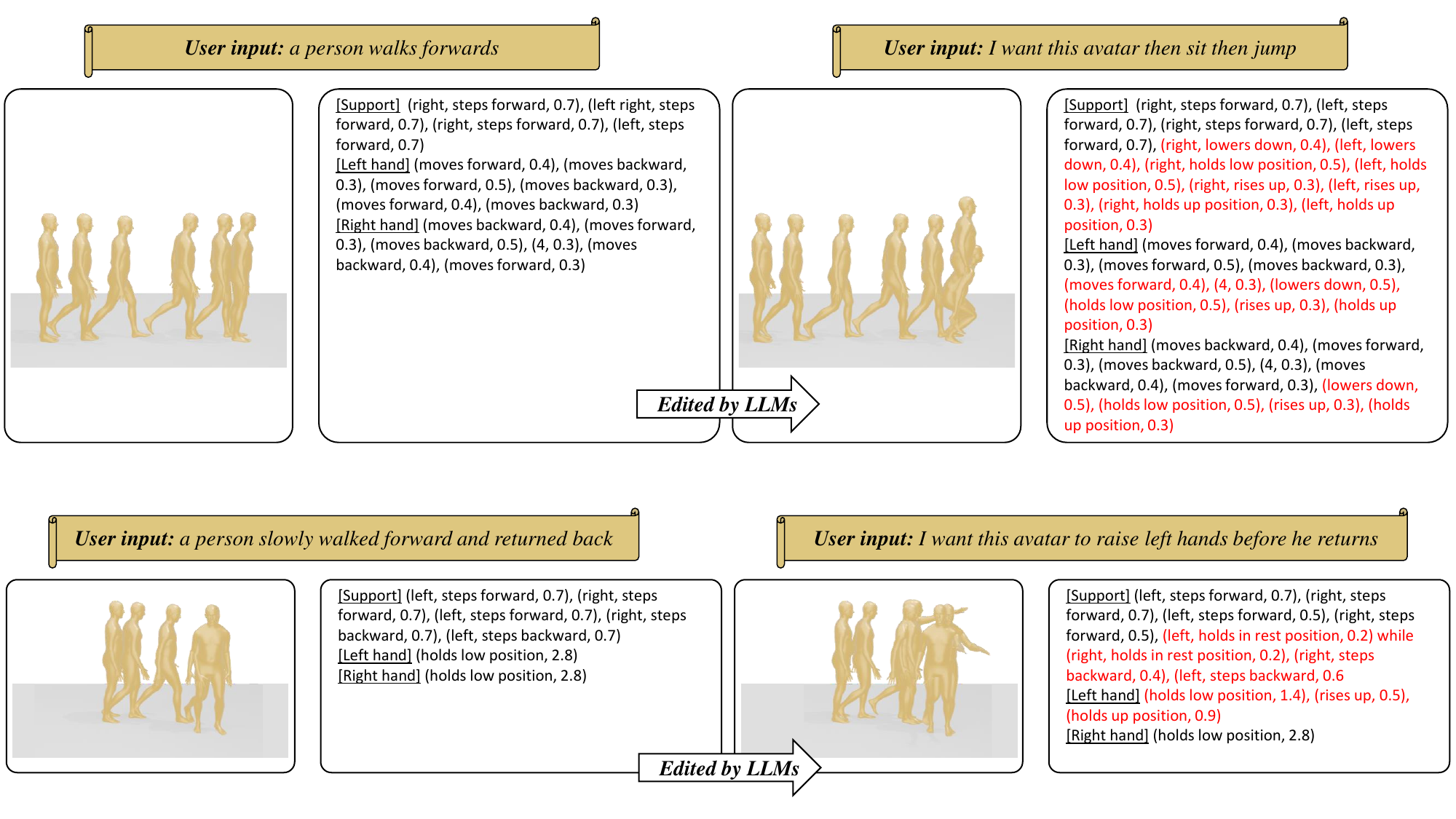}
\caption{Visual examples for motion editing, with LLM-composed description. Left: the initial generated motion conditioned by the user input. Right: an ambiguous edit instruction is given to modify the motion. The corresponding LLM-composed description follows the LabanLite-defined format ``\texttt{<body-part group>, <moving semantic>, <time>}''. We highlight the refined part with red to show LLMs' dialogue context understanding capability.}
\label{fig:sup_mot_edit}
\end{figure*}

\subsubsection{Visual Examples of Motion Editing}

We demonstrate the potential of our framework in terms of motion editing capability. Fig.~\ref{fig:sup_mot_edit} presents examples of edited motions alongside the corresponding conceptual descriptions refined by the LLM. The refined part is also highlighted. As shown, the powerful ability of LLMs to interpret ambiguous instructions and understand dialogue context enables our framework to support interactive motion modification through natural language conversation. This capability, which allows users to intuitively edit motions during dialogue, represents a significant advancement over existing works.

\subsubsection{Additional Comparison Examples}

\begin{figure*}[t]
\centering
\includegraphics[width=0.75\linewidth]{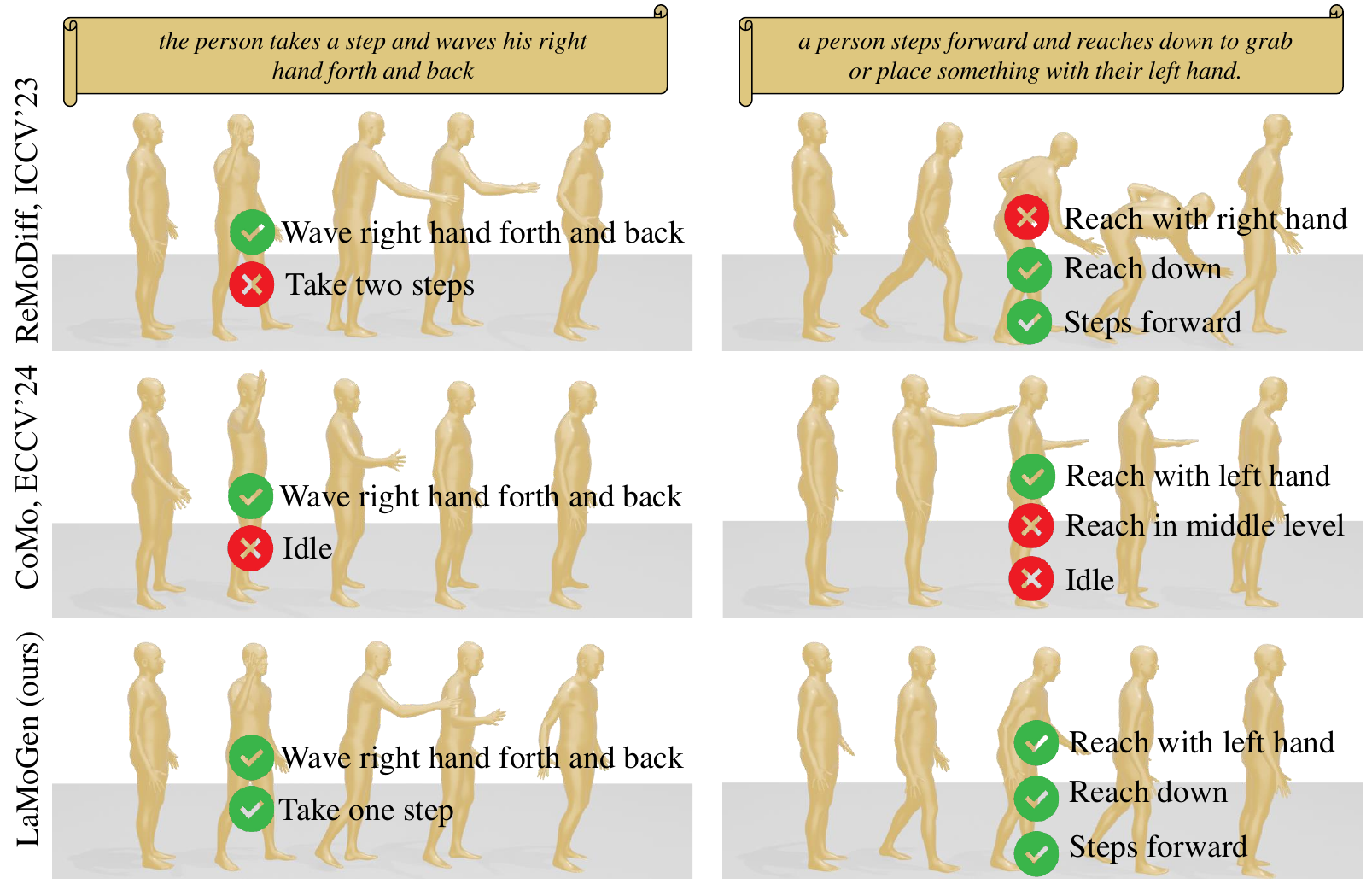}
\caption{Qualitative comparisons on HumanML3D test sets, with motions progressing from left to right. Misalignments between text and generated motions are highlighted.}
\label{fig:sup_quali}
\end{figure*}

Additional comparison examples are shown in Fig.~\ref{fig:sup_quali}. From this figure, we can observe that ReMoDiff and CoMo occasionally violate chronological order and struggle to specify movement timings, while our method generates text-aligned motions.
Comprehensive video results, including both successful examples and failure cases, are provided in the supplementary materials. These video results were also utilised in the user study.

\begin{table}[t]
\setlength{\tabcolsep}{0.5mm}
\centering\small
\caption{Ablation study of different numbers of top-matched references and masking ratios on the HumanML3D test set.}
\begin{tabular}{rcccc}
\toprule
\textbf{Top ref.} & R@3 $\uparrow$ & FID $\downarrow$ & MM-Dist $\downarrow$ & Multi-Mod. $\uparrow$ \\ 
\midrule
1 & 0.755 & 0.595 & 3.840 & 0.954 \\
\rowcolor{gray!10}
3 & 0.796 & 0.252 & 3.087 & 1.131 \\ 
5 & 0.782 & 0.371 & 3.191 & 1.030 \\
7 & 0.774 & 0.324 & 3.225 & 0.971 \\
\midrule
\textbf{Mask rat.} & R@3 $\uparrow$ & FID $\downarrow$ & MM-Dist $\downarrow$ & Multi-Mod. $\uparrow$ \\ 
\midrule
0.1 & 0.789 & 0.258 & 3.008 & 1.092 \\ 
\rowcolor{gray!10}
0.3 & 0.796 & 0.252 & 3.087 & 1.131 \\ 
0.5 & 0.787 & 0.299 & 3.211 & 1.011 \\
0.7 & 0.772 & 0.345 & 3.290 & 0.992 \\
0.9 & 0.705 & 0.457 & 3.302 & 0.985 \\
\bottomrule
\end{tabular}
\label{tab:ab}
\end{table}

\begin{table}[t]
\setlength{\tabcolsep}{3mm}
\centering\small
\caption{Ablation study of different code dimensions on the HumanML3D test set. AITS measures the average inference time cost.}
\begin{tabular}{rccc}
\toprule
\textbf{Code Dim} & R@3 $\uparrow$ & FID $\downarrow$ & AITS $\downarrow$  \\ 
\midrule
CoMo~\citep{huang2024controllable} & 0.790 & 0.262 & 0.58 \\
\midrule
256 & 0.759 & 0.310 & 0.28  \\
384 & 0.783 & 0.295 & 0.32  \\
\rowcolor{gray!10}
512 & 0.796 & 0.252 & 0.45  \\ 
768 & 0.802 & 0.241 & 0.57  \\
1024 & 0.804 & 0.227 & 0.65  \\
\bottomrule
\end{tabular}
\label{tab:ab_code_dim}
\end{table}

{
\setlength{\tabcolsep}{1mm}
\begin{table}[t]
\centering\small
\caption{Evaluation on Laban Benchmark. $\dagger$: main paper setting.}
\scalebox{0.83}{
\begin{tabular}{rccccc}
\toprule
Method & avg. SMT $\uparrow$ & avg. TMP $\uparrow$ & avg. HMN $\uparrow$ & R@3 $\uparrow$ & FID $\downarrow$ \\
\midrule
Real data & - & - & - & 0.216 & 0.001 \\
\cmidrule(lr){1-6}
MDM & 0.338 & 0.298 & 0.201 & 0.180 & 22.81 \\
MDM$_{lbn}$ & 0.475 & 0.438 & 0.271 & 0.204 & 1.932 \\
GPT4.1 $\dagger$ & 0.533 & 0.501 & 0.392 & 0.208 & 1.861 \\
GPT4.1$_{ip}$ & 0.496 & 0.467 & 0.328 & 0.200 & 3.408 \\
\bottomrule
\end{tabular}
}
\label{tab:rebuttal_loco}
\end{table}
}

{
\setlength{\tabcolsep}{0.5mm}
\begin{table}[t]
\centering\small
\caption{Evaluation on HumanML3D. We examine the impact of varying the number of evenly spaced intervals for the Direction and Level symbols. $\dagger$: main paper setting. }
\scalebox{0.83}{
\begin{tabular}{rccccccc}
\toprule
\multicolumn{2}{r}{Method} & R@3 $\uparrow$ & FID $\downarrow$ & MM-Dist $\downarrow$ & Div. $\rightarrow$ & Multi-Mod. $\uparrow$ & Token Cost (k) $\downarrow$ \\
\midrule
\multicolumn{2}{r}{Real data} & 0.797 & 0.002 & 2.974 & 9.503 & - & - \\
\cmidrule(lr){1-8}
\multicolumn{2}{r}{MDM} & 0.611 & 0.544 & 5.566 & 9.559 & 2.799 & - \\
\multicolumn{2}{r}{CoMo} & 0.790 & 0.262 & 3.032 & 9.936 & 1.013 & 6.5 \\
\multirow{3}{*}{\rotatebox{90}{\shortstack{Interval}}} & 3 $\dagger$ & 0.796 & 0.252 & 3.087 & 9.124 & 1.131 & 5.0 \\
& 5 & 0.796 & 0.198 & 3.026 & 9.859 & 1.360 & 7.7 \\
& 7 & 0.800 & 0.184 & 2.992 & 9.840 & 1.413 & 10.3 \\
\multicolumn{2}{r}{BERT} & 0.801 & 0.199 & 3.001 & 3.409 & 0.943 & 5.0 \\
\multicolumn{2}{r}{Qwen3-8b} & 0.770 & 0.537 & 3.849 & 10.432 & 1.205 & 5.0 \\
\bottomrule
\end{tabular}
}
\label{tab:rebuttal_humanml3d}
\end{table}
}

\subsection{Ablation Studies}

Due to page limitations, the numerical results of our ablation studies are included in the supplementary material, while analyses and conclusions are presented in the main paper. Below, we reiterate the ablation settings together with the corresponding quantitative findings. Additionally, we report the impact of the dimension of codebook entries.

\paragraph{Number of top-matched references.}
For LLM-guided conceptual composition, we provide top-matched conceptual examples to guide the composition of new conceptual codes via LLMs' retrieval augmentation prompting. We investigate how the number of retrieved examples affects performance on the HumanML3D test set using GPT-4.1. As shown in Table~\ref{tab:ab}, increasing the number of examples from 1 to 3 consistently improves performance, indicating that LLMs benefit from sufficient references for accurate imitation. However, when increasing the number of examples further (e.g., from 3 to 5 or 7), we do not observe further improvement. This phenomenon is mainly caused by the limitation of the model's context window --- the maximum number of tokens (words, symbols, etc.) that the LLM can process at once. When too many examples are provided, the beginning of the context may be truncated or, even within the window limit, earlier examples may be forgotten due to the model's recency bias. As a result, LLMs tend to focus more on the most recently presented cues and ignore information from previous examples, thereby diminishing the potential benefit of providing more reference cases.

\paragraph{Masking ratio on Laban codes.}
We also examine how the masking ratio on Laban codes affects performance during code generation. A higher masking ratio reduces the influence of conceptual cues, making motion generation rely more on text and less on Laban symbols, reducing the influence of conceptual guidance. We conduct this study on the HumanML3D test set, with GPT4.1 and top-3 retrieval settings. Comparing results in Table~\ref{tab:ab}, a masking ratio of 0.3 offers the best balance and achieves optimal performance.

\paragraph{Impact of Code Dimension.}
We further investigate the impact of varying dimensions of the Laban code on the HumanML3D test set. Following the ablation findings above, the number of retrieved references is set to 3, and the masking ratio to 0.3, with GPT-4.1 as the LLM. 
We assess the generation quality (R@3, FID) and Average Inference Time per Sentence (AITS) across different code dimensions, and compare our approach to CoMo~\citep{huang2024controllable}. 
As shown in Table~\ref{tab:ab_code_dim}, increasing the code dimension improves generation performance but also increases inference time. Considering both factors, a code dimension of 512 is selected as the optimal setting.

\paragraph{Impact of Symbolic Guidance.}
To investigate the effect of symbolic guidance, we integrate LLM-composed symbolic codes into the baseline MDM model~\cite{tevet2023mdm} by summing the activated code with the initial noise through the diffusion process. As reported in Table.~\ref{tab:rebuttal_loco} (``MDM$_{lbn}$''), this symbolic conditioning results in a substantial improvement: FID is reduced by a factor of 20 compared to the vanilla MDM. This dramatic gain demonstrates that leveraging high-level symbolic structure from language models can provide strong guidance for motion generation, significantly enhancing both overall fidelity and alignment between text and generated motion.

\paragraph{Impact on Instruction Prompting.}
Our LLM composition process relies on two components: Retrieval-Augmented Generation (RAG), which supplies templates for sub-action representations and temporal durations, and Instruction Prompting (IP), which defines the rules and definition on Conceptual Symbols. To evaluate the specific contribution of RAG, we conduct an ablation in which RAG is removed and the LLM is tasked with composing concepts solely based on IP instructions.  Table.~\ref{tab:rebuttal_loco} ``GPT4.1$_{ip}$'' shows that this configuration results in a slight drop across most evaluation metrics. Importantly, we observe a much larger FID gap; without the concrete sub-action durations provided by RAG, the generated duration distributions diverge significantly from ground truth. 

\paragraph{Impact of Discrete Intervals.} 
To further examine the effect of discretization granularity in LabanLite, we conducted an ablation study varying the number of discrete intervals in our encoding scheme. As shown in Table.~\ref{tab:rebuttal_humanml3d}, the ``Interval-7'' configuration improves FID by $26\%$, but also doubles the LLM cost. It shows that, increasing the number of intervals can improve the reconstruction fidelity, but poses challenges for computational cost.  Moreover, excessive granularity necessitates the introduction of nonstandard symbols, which diminishes the interpretability for human readers trying to understand the notation. Overall, our results suggest that a moderate number of intervals yields a favorable balance between fidelity, interpretability, and computational efficiency.

\paragraph{Choice of GPT-style vs. BERT-style Generation.}
In our framework, the availability of global information allows for both GPT-style (autoregressive) and BERT-style (infilling) generation strategies. As observed in Table.~\ref{tab:rebuttal_humanml3d}, adopting BERT-style generation (``BERT'') improves the FID metric by 21\%, reflecting enhanced accuracy in reconstruction. However, this comes at the expense of diversity, which decreases significantly by 62.6\%. In contrast, GPT-style generation (``3 $\dagger$'') offers superior diversity, which is crucial for the downstream requirements of our task. Balancing these factors, we prioritize generation diversity and thus adopt GPT's autoregressive approach as our default method.

\paragraph{Performance on Weaker Language Models.}
To evaluate the robustness of our method with less powerful language models, we assess performance using the Qwen3-8b model, as reported in Table.~\ref{tab:rebuttal_humanml3d}. As expected, reducing LLM size to 8B leads to slightly lower overall generation metrics. Nevertheless, by incorporating RAG and IP constraints, our method successfully maintains valid Laban symbols, ensuring strong alignment between textual inputs and motion outputs. Notably, the Qwen3-8b configuration achieves a $26\%$ improvement in R@3 compared to the MDM baseline. These results indicate our approach remains effective—even when deployed with weaker LLMs—highlighting its versatility for resource-constrained scenarios.

\section{Discussion}

To address the challenge posed by ultra-long text inputs (e.g., ``walk slowly for 10 steps, then jump twice while waving both hands''), we can employ the LLM itself to segment complex user instructions into manageable sub-actions (e.g., ``walk slowly for 10 steps'' and ``jump twice while waving both hands''). This decomposition is well within the capabilities of modern LLMs, and crucially, the semantic integrity of the original motion intent is preserved during splitting.
Once the ultra-long instruction is divided into discrete sub-actions, each segment can be processed independently, enabling LaMoGen to handle extended or sequential behaviors in a scalable manner. The concatenation of generated sub-actions naturally forms an ultra-long motion sequence, overcoming the constraints imposed by limited LLM context windows.

\section{Limitations and Future Work}\label{sec:sup_plan}

\subsection{Need for a User Interface}
Due to constraints in research funding and computational resources, the development of a user interface for editing Laban symbol sequences was beyond the scope of the present work. Nevertheless, we recognise that such an interface would be highly beneficial for promoting the broader adoption of the proposed LaMoGen framework and the LabanLite motion representation. As part of our future work, we plan to design and implement a user-friendly LaMoGen interface, which could be integrated as an add-on for 3D creation tools such as Blender.

\subsection{Learning Cost of Labanotation}
Learning Labanotation does entail a certain learning cost. While incorporating a larger number of Laban symbols allows for the expression of more complex motions, it also inevitably introduces more intricate notation rules. In the main paper, we address this challenge by proposing a {two-stage LLM-Guided Text-Laban-Motion Generation Module}: in the first stage, either human users or LLMs compose conceptual Laban instances to represent high-level motion; in the second stage, a Kinematic Detail Augmentor further refines these conceptual instances into detailed low-level motion. This two-stage approach simplifies the notation by requiring only conceptual symbols—approximately one-fifth of the full symbol set—and thus helps to reduce the learning effort. Although a certain amount of learning is still necessary, we believe that our approach is more accessible and convenient for human editing compared to existing methods based on key frames~\citep{liu2023plan,wang2025stickmotion,au2025soscontrol} or key points~\citep{wan2024tlcontrol}.

\subsection{Potential for Disentangling Motion and Style}
It is important to note that Labanotation applies quantization simultaneously at both the spatial and temporal levels. As a result, distinct performances—specifically, different individuals' personalized interpretations of the same contextual prompt—may be encoded using identical Laban symbols. This inherent abstraction implies that LabanLite, which is built upon Labanotation, does not explicitly represent fine-grained individual differences.

For instance, consider the context of ``raising the hand.'' Younger individuals may execute this action energetically, characterized by a rapid initial lift followed by a deceleration as the hand approaches and stabilizes at the target height. In contrast, older individuals may perform the same movement with a composed steadiness, gradually lifting the hand at a constant speed from a lower to a higher position. These two performances exhibit distinctly different profiles in terms of acceleration and deceleration, with the former manifesting a vigorous style and the latter reflecting a calm, controlled demeanor. Notably, despite these pronounced differences in movement dynamics—each indicative of a unique personal style—the resulting motion sequences are encoded identically within the Laban symbol sequence.

Rather than viewing this as a limitation, we recognize it as an opportunity: by abstracting away such personalized nuances, LabanLite naturally separates the essential motion content from individual stylistic expression. Therefore, LabanLite holds significant potential for disentangling motion and style. In future work, we aim to further investigate this property by treating LabanLite-based motion descriptions as representations of neutral, canonical movements and subsequently modeling style-specific residuals to enable flexible motion restyling.

\subsection{Potential for Flexible Extension Across Body Parts and Domains}

The proposed LabanLite representation and LaMoGen framework possess significant flexibility in modeling human movement. While our current implementation focuses on the major body parts, LabanLite’s codebook and design can be readily extended to model additional components, such as finger and toe gestures, simply by incorporating the corresponding Laban symbols. This adaptability is supported by recent studies, which have demonstrated the feasibility of Laban-based fine-grained motion reconstruction in sub-domains such as finger gesture estimation~\citep{li2024translating} and dance movement reconstruction~\citep{li2023labanformer}.
Therefore, although our present work primarily targets daily motion, the methodology can be straightforwardly generalized to facilitate applications in a wide range of domains, including detailed human–object interactions and expressive dance analysis. We anticipate that future work will broaden the scope of LabanLite, establishing a unified motion representation applicable to diverse areas of human movement science and interactive systems.

\begin{table*}[t]
\centering
\caption{Prompts used in our LaMoGen to process natural language via Large Language Models.}
\begin{tabularx}{0.8\textwidth}{X}
\toprule
\rowcolor{blue!10}
Prompt \#1: Conceptual cues composer \\
\midrule
There are five digit collections describing movements, where each line consists of: [number] [Caption] - a general description of the motion sequence. [Support] - detailed descriptions of the movements of the supporting body parts, specifically the left and right feet, using a series of triplets. [Left hand] - detailed descriptions of the movements of the left hand, using a series of tuples. [Right hand] - detailed descriptions of the movements of the right hand, using a series of tuples. In the detailed descriptions, we specify the movement details for each body part and their duration in seconds. For the support movements, the details must be selected from these 54 categories: [1: steps to rest position, ..., 54: holds in knee-flexed backward diagonally to left position]. For the hand movements, the details must be selected from these 81 categories: [1: moves close to shoulder, ..., 81: moves relatively low backward diagonally to left]. For example, for the [Support] line, the triplet list would be like: (left, 1, 0.25), (right, 2, 0.25), (left, 1, 0.25) while (right, 2, 0.25). This means that the first movement is ``left foot steps to rest position in 0.25 seconds''. The second movement is ``right foot steps forward in 0.25 seconds''. The third movement is ``left foot steps to rest position in 0.25 seconds while right foot steps forward in 0.25 seconds''. For the [Left hand] line, the tuple list would be like: (1, 0.5), (2, 0.2). This means that the first movement is ``left hand moves close to shoulder in 0.5 seconds'' and the second movement is ``left hand moves forward in 0.2 seconds''. For the [Right hand] line, the structure and definition are similar to [Left hand] lines. Below is the main body of the digit collection describing the movements. You should strictly imitate the following content and create only one digit collection of \textit{YOUR\_INPUT}. Reply without explanation. \\
\midrule
\rowcolor{blue!10}
Prompt \#2: Rephraser for generating motion descriptions \\
\midrule
Rephrase the sentence creatively \textit{YOUR\_INPUT}. the step number is \textit{YOUR\_INPUT}, with the step order: \textit{YOUR\_INPUT}. \\
\bottomrule
\end{tabularx}
\label{tab:prompt}
\end{table*} 

{
\setlength{\tabcolsep}{1mm}
\begin{table*}[t]
\centering
\caption{Illustration of the Direction symbols and their corresponding partial semantics.}
\begin{tabular}{cm{1cm}c}
\toprule
Name & Appearance & Semantics \\
\midrule
Direction L/F & \centering\includegraphics[width=0.6cm]{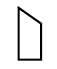} & Move body part left forward\\
Direction M/F & \centering\includegraphics[width=0.6cm]{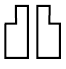} & Move body part to front \\
Direction R/F & \centering\includegraphics[width=0.6cm]{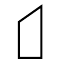} & Move body part right forward\\
Direction L/M & \centering\includegraphics[width=0.6cm]{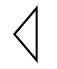} & Move body part to left\\
Direction M/M & \centering\includegraphics[width=0.6cm]{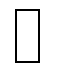} & Move body part to middle\\
Direction R/M & \centering\includegraphics[width=0.6cm]{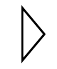} & Move body part to right\\
Direction L/B & \centering\includegraphics[width=0.6cm]{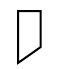} & Move body part left backward\\
Direction M/B & \centering\includegraphics[width=0.6cm]{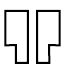} & Move body part to back\\
Direction R/B & \centering\includegraphics[width=0.6cm]{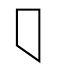} & Move body part right backward\\
\bottomrule
\end{tabular}
\label{tab:direction}
\end{table*}
}

{
\setlength{\tabcolsep}{1mm}
\begin{table*}[t]
\centering
\caption{Illustration of the Level symbols and their corresponding partial semantics.}
\begin{tabular}{cm{1cm}c}
\toprule
Name & Appearance & Semantics \\
\midrule
Level Hi. & \centering\includegraphics[width=0.6cm]{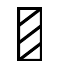} & Move body part to high level\\
Level Mi. & \centering\includegraphics[width=0.6cm]{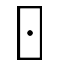} & Move body part to mid-level \\
Level Lo. & \centering\includegraphics[width=0.6cm]{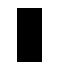} & Move body part to low level\\
\bottomrule
\end{tabular}
\label{tab:level}
\end{table*}
}

{
\setlength{\tabcolsep}{1mm}
\begin{table*}[t]
\centering
\caption{Illustration of the Hold symbols and their corresponding partial semantics. Note that if the attribute field for a symbol is left empty, it indicates that this body part is dynamic.}
\begin{tabular}{cm{1cm}c}
\toprule
Name & Appearance & Semantics \\
\midrule
Hold & \centering\includegraphics[width=0.5cm]{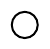} & Body part is stationary\\
\bottomrule
\end{tabular}
\label{tab:hold}
\end{table*}
}

{
\setlength{\tabcolsep}{1mm}
\begin{table*}[t]
\centering
\caption{Illustration of the Orientation symbols and their corresponding partial semantics. Note that according to Labanotation, each symbol does not have a specific name. In LabanLite, we simply assign them sequential names for convenience.}
\begin{tabular}{cm{1cm}c}
\toprule
Name & Appearance & Semantics \\
\midrule
Orient 0 & \centering\includegraphics[width=0.6cm]{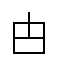} & Body part orients at around $0^\circ$\\
Orient 1 & \centering\includegraphics[width=0.6cm]{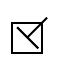} & Body part orients at around $45^\circ$\\
Orient 2 & \centering\includegraphics[width=0.6cm]{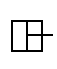} & Body part orients at around $90^\circ$\\
Orient 3 & \centering\includegraphics[width=0.6cm]{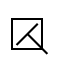} & Body part orients at around $135^\circ$\\
Orient 4 & \centering\includegraphics[width=0.6cm]{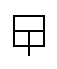} & Body part orients at around $180^\circ$\\
Orient 5 & \centering\includegraphics[width=0.6cm]{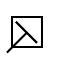} & Body part orients at around $225^\circ$\\
Orient 6 & \centering\includegraphics[width=0.6cm]{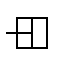} & Body part orients at around $270^\circ$\\
Orient 7 & \centering\includegraphics[width=0.6cm]{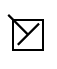} & Body part orients at around $315^\circ$\\
\bottomrule
\end{tabular}
\label{tab:orient}
\end{table*}
}

{
\setlength{\tabcolsep}{1mm}
\begin{table*}[t]
\centering
\caption{Illustration of the Bend symbols and their corresponding partial semantics. Note that according to Labanotation, each symbol does not have a specific name. In LabanLite, we simply assign them sequential names for convenience.}
\begin{tabular}{cm{1cm}c}
\toprule
Name & Appearance & Semantics \\
\midrule
Bend 0 & \centering\includegraphics[width=0.6cm]{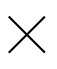} & Body part bends at around $0^\circ$\\
Bend 1 & \centering\includegraphics[width=0.6cm]{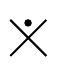} & Body part bends at around $30^\circ$\\
Bend 2 & \centering\includegraphics[width=0.6cm]{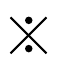} & Body part bends at around $60^\circ$\\
Bend 3 & \centering\includegraphics[width=0.6cm]{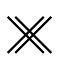} & Body part bends at around $90^\circ$\\
Bend 4 & \centering\includegraphics[width=0.6cm]{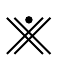} & Body part bends at around $120^\circ$\\
Bend 5 & \centering\includegraphics[width=0.6cm]{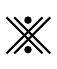} & Body part bends at around $150^\circ$\\
\bottomrule
\end{tabular}
\label{tab:bend}
\end{table*}
}

{
\setlength{\tabcolsep}{1mm}
\begin{table*}[t]
\centering
\caption{Illustration of the Moving-effort symbols and their corresponding partial semantics. Note that according to Labanotation, each symbol does not have a specific name. In LabanLite, we simply assign them sequential names for convenience. If the attribute field for a symbol is left empty, it indicates ``Moving-effort 1''.}
\begin{tabular}{cm{1cm}c}
\toprule
Name & Appearance & Semantics \\
\midrule
Moving-effort 0 & \centering\includegraphics[width=0.6cm]{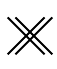} & Body part moves very slow\\
Moving-effort 1 & \centering None & Body part moves in normal speed \\
Moving-effort 2 &  \centering\includegraphics[width=0.6cm]{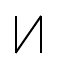} & Body part moves fast \\
Moving-effort 3 &  \centering\includegraphics[width=0.6cm]{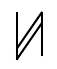} & Body part moves very fast \\
\bottomrule
\end{tabular}
\label{tab:speed}
\end{table*}
}

\begin{table*}[t]
\centering\small
\caption{Definition of the Laban codebook. Each codebook entry is described by its index (Code Idx.), associated Body-Part Group, corresponding SMPL key joint name (SMPL Joint), relevant attribute (Attribute), a marker indicating whether the attribute is conceptual (is Concpt.), and the Laban staff column name (Staff Col.). }
\begin{tabularx}{1\textwidth}{p{1.5cm}p{1.5cm}XXp{1cm}p{2.2cm}}
\toprule 
Code Idx. & Body-Part Group & SMPL Joint & Attribute & is Concpt. & Staff Col. \\
\midrule
$1\sim3$ & \multirow{6}{*}{Support-L} & {Left foot} & Direction (L/M/R) & \cmark & {Left support} \\
$4\sim6$ & & {Left foot} & Direction (B/M/F) & \cmark & {Left support} \\
$7\sim9$ & & {Left foot} & Level (Lo./Mi./Hi.) & \cmark & {Left support} \\
$10\sim15$ & & Left knee & Bend & \xmark & Left leg gesture \\ 
$16\sim21$ & & Left hip & Bend & \xmark & Left leg gesture \\ 
$22\sim23$ & & Left foot \& knee \& hip & Hold & \cmark & {Left support} \\
\midrule
$23\sim26$ & \multirow{6}{*}{Support-R} & Right foot & Direction (L/M/R) & \cmark & Right support \\
$27\sim29$ & & Right foot & Direction (B/M/F) & \cmark & Right support \\
$30\sim32$ & & Right foot & Level (Lo./Mi./Hi.) & \cmark & Right support \\
$33\sim38$ & & Right knee & Bend & \xmark & Right leg gesture \\ 
$39\sim44$ & & Right hip & Bend & \xmark & Right leg gesture \\ 
$45\sim46$ & & Right foot \& knee \& hip & Hold & \cmark & Right support \\
\midrule
$47\sim54$ & \multirow{4}{*}{Support-Both} & Pelvis & Orient. Horiz. & \xmark & Body (Whole) \\
$55\sim62$ & & Pelvis & Orient. Vert. & \xmark & Body (Whole) \\
$63\sim67$ & & Pelvis & Moving effort Horiz. & \xmark & Body (Whole) \\
$68\sim72$ & & Pelvis & Moving effort Vert. & \xmark & Body (Whole) \\
\midrule
$73\sim75$ & \multirow{6}{*}{Upper-L} & Left hand & Direction (L/M/R) & \cmark & Left hand \\
$76\sim78$ & & Left hand & Direction (B/M/F) & \cmark & Left hand \\
$79\sim81$ & & Left hand & Level (Lo./Mi./Hi.) & \cmark & Left hand \\
$82\sim87$ & & Left elbow & Elbow Bend & \xmark & Left arm \\ 
$88\sim93$ & & Left shoulder & Shoulder Bend & \xmark & Left arm \\ 
$94\sim95$ & & Left hand \& elbow \& shoulder & Hold & \cmark & Left hand \\
\midrule
$96\sim98$ & \multirow{6}{*}{Upper-R} & Right hand & Direction (L/M/R) & \cmark & Right hand \\
$99\sim101$ & & Right hand & Direction (B/M/F) & \cmark & Right hand \\
$102\sim104$ & & Right hand & Level (Lo./Mi./Hi.) & \cmark & Right hand \\
$105\sim110$ & & Right elbow & Elbow Bend & \xmark & Right arm \\ 
$111\sim116$ & & Right shoulder & Shoulder Bend & \xmark & Right arm \\ 
$117\sim118$ & & Right hand \& elbow \& shoulder & Hold & \cmark & Right hand \\
\midrule
$119\sim126$ & \multirow{3}{*}{Torso} & Head & Orient. Horiz. & \xmark & Head \\ 
$127\sim134$ & & Head & Orient. Vert. & \xmark & Head \\ 
$135\sim140$ & & Spine2 & Bend & \xmark & Body (Whole) \\
\midrule
$141\sim143$ & \multirow{3}{*}{Upper-L} & Left elbow & Direction (L/M/R) & \xmark & Left arm \\
$144\sim146$ & & Left elbow & Direction (B/M/F) & \xmark & Left arm \\
$147\sim149$ & & Left elbow & Level (Lo./Mi./Hi.) & \xmark & Left arm \\
\midrule
$150\sim152$ & \multirow{3}{*}{Upper-R} & Right elbow & Direction (L/M/R) & \xmark & Right arm \\
$153\sim155$ & & Right elbow & Direction (B/M/F) & \xmark & Right arm \\
$156\sim158$ & & Right elbow & Level (Lo./Mi./Hi.) & \xmark & Right arm \\
\bottomrule
\end{tabularx}
\label{tab:codebook}
\end{table*}

\begin{table*}[t]
\centering\small
\caption{Support body-part's moving semantic lookup table.}
\begin{tabularx}{0.95\textwidth}{p{0.45cm}p{4.8cm}p{0.45cm}X}
\toprule 
Index & Semantics & Index & Semantics \\
\midrule
1 &  steps to rest position & 28 &  holds in rest position \\
2 &  steps forward & 29 &  holds in forward position \\
3 &  steps backward & 30 &  holds in backward position \\
4 &  steps to right & 31 &  holds in right position \\
5 &  steps to left & 32 &  holds in left position \\
6 &  steps forward diagonally to right & 33 &  holds in forward diagonally to right position \\
7 &  steps forward diagonally to left & 34 &  holds in forward diagonally to left position \\
8 &  steps backward diagonally to right & 35 &  holds in backward diagonally to right position \\
9 &  steps backward diagonally to left & 36 &  holds in backward diagonally to left position \\
10 &  rises & 37 &  holds in the raised position \\
11 &  rises to forward & 38 &  holds in the raised forward position \\
12 &  rises to backward & 39 &  holds in the raised backward position \\
13 &  rises to right & 40 &  holds in the raised right position \\
14 &  rises to left & 41 &  holds in the raised left position \\
15 &  rises forward diagonally to right & 42 &  holds in the raised forward diagonally to right position \\
16 &  rises forward diagonally to left & 43 &  holds in the raised forward diagonally to left position \\
17 &  rises backward diagonally to right & 44 &  holds in the raised backward diagonally to right position \\
18 &  rises backward diagonally to left & 45 &  holds in the raised backward diagonally to left position \\
19 &  knee flex & 46 &  holds in knee-flexed position \\
20 &  knee flex forward & 47 &  holds in knee-flexed forward position \\
21 &  knee flex backward & 48 &  holds in knee-flexed backward position \\
22 &  knee flex right & 49 &  holds in knee-flexed right position \\
23 &  knee flex left & 50 &  holds in knee-flexed left position \\
24 &  knee flex forward diagonally to right & 51 &  holds in knee-flexed forward diagonally to right position \\
25 &  knee flex forward diagonally to left & 52 &  holds in knee-flexed forward diagonally to left position \\
26 &  knee flex backward diagonally to right & 53 &  holds in knee-flexed backward diagonally to right position \\
27 &  knee flex backward diagonally to left & 54 &  holds in knee-flexed backward diagonally to left position \\
\bottomrule
\end{tabularx}
\label{tab:sup_semantic}
\end{table*}

\begin{table*}[t]
\centering\small
\caption{Arm body-part's moving semantic lookup table.}
\begin{tabularx}{0.95\textwidth}{p{0.45cm}Xp{0.45cm}Xp{0.45cm}X}
\toprule
Index & Semantics & Index & Semantics & Index & Semantics \\
\midrule
1 &  moves close to shoulder & 28 &  holds close to shoulder position & 55 &  moves relatively to previous position \\
2 &  moves forward & 29 &  holds forward position & 56 &  moves relatively forward \\
3 &  moves backward & 30 &  holds backward position & 57 &  moves relatively backward \\
4 &  moves to right & 31 &  holds right position & 58 &  moves to relatively right \\
5 &  moves to left & 32 &  holds left position & 59 &  moves to relatively left \\
6 &  moves forward diagonally to right & 33 &  holds forward diagonally to right position & 60 &  moves relatively forward diagonally to right \\
7 &  moves forward diagonally to left & 34 &  holds forward diagonally to left position & 61 &  moves relatively forward diagonally to left \\
8 &  moves backward diagonally to right & 35 &  holds backward diagonally to right position & 62 &  moves relatively backward diagonally to right \\
9 &  moves backward diagonally to left & 36 &  holds backward diagonally to left position & 63 &  moves relatively backward diagonally to left \\
10 &  rises up & 37 &  holds up position & 64 &  moves relatively up \\
11 &  rises to up forward & 38 &  holds up forward position & 65 &  moves relatively up forward \\
12 &  rises to up backward & 39 &  holds up backward position & 66 &  moves relatively up backward \\
13 &  rises to up right & 40 &  holds up right position & 67 &  moves relatively up right \\
14 &  rises to up left & 41 &  holds up left position & 68 &  moves relatively up left \\
15 &  rises up forward diagonally to right & 42 &  holds up forward diagonally to right position & 69 &  moves relatively up forward diagonally to right \\
16 &  rises up forward diagonally to left & 43 &  holds up forward diagonally to left position & 70 &  moves relatively up forward diagonally to left \\
17 &  rises up backward diagonally to right & 44 &  holds up backward diagonally to right position & 71 &  moves relatively up backward diagonally to right \\
18 &  rises up backward diagonally to left & 45 &  holds up backward diagonally to left position & 72 &  moves relatively up backward diagonally to left \\
19 &  lowers down & 46 &  holds low position & 73 &  moves relatively low \\
20 &  lowers to down forward & 47 &  holds low forward position & 74 &  moves relatively low forward \\
21 &  lowers to down backward & 48 &  holds low backward position & 75 &  moves relatively low backward \\
22 &  lowers to down right & 49 &  holds low right position & 76 &  moves relatively low right \\
23 &  lowers to down left & 50 &  holds low left position & 77 &  moves relatively low left \\
24 &  lowers down forward diagonally to right & 51 &  holds low forward diagonally to right position & 78 &  moves relatively low forward diagonally to right \\
25 &  lowers down forward diagonally to left & 52 &  holds low forward diagonally to left position & 79 &  moves relatively low forward diagonally to left \\
26 &  lowers down backward diagonally to right & 53 &  holds low backward diagonally to right position & 80 &  moves relatively low backward diagonally to right \\
27 &  lowers down backward diagonally to left & 54 &  holds low backward diagonally to left position & 81 &  moves relatively low backward diagonally to left \\
\bottomrule
\end{tabularx}
\label{tab:arm_semantic}
\end{table*}

\end{document}